\documentclass[]{fairmeta}

\usepackage{amssymb}%
\usepackage{pifont}%

\usepackage{symbols}

\usepackage{amsmath,amsfonts,bm}

\def\1{\bm{1}}

\def\sp{space}

\DeclareMathAlphabet{\mathsfit}{\encodingdefault}{\sfdefault}{m}{sl}
\SetMathAlphabet{\mathsfit}{bold}{\encodingdefault}{\sfdefault}{bx}{n}

\newcommand*{\ShowNotes}{} %
\usepackage{soul}
\usepackage{multirow}

\definecolor{lightred}{rgb}{0.9,0.4,0.4}
\definecolor{darkred}{rgb}{0.7,0.1,0.1}
\definecolor{darkgreen}{rgb}{0.1,0.7,0.1}
\definecolor{cyan}{rgb}{0.7,0.0,0.7}
\definecolor{dblue}{rgb}{0.2,0.2,0.8}
\definecolor{maroon}{rgb}{0.76,.13,.28}
\definecolor{burntorange}{rgb}{0.81,.33,0}
\definecolor{tealblue}{rgb}{0.212,0.459, 0.533}
\definecolor{myyellow}{rgb}{0.8627451 , 0.75294118, 0.20784314]}
\definecolor{mypink}{rgb}{0.93359375, 0.62109375, 0.83984375}
\definecolor{pp}{rgb}{0.43921569, 0.18823529, 0.62745098}
\definecolor{rr}{rgb}{0.5254902 , 0.00784314, 0.12941176}
\definecolor{bb}{rgb}{0.09019608, 0.23529412, 0.37647059}
\definecolor{yy}{rgb}{0.49803922, 0.3372549 , 0.0}
\definecolor{gg}{rgb}{0.02352941, 0.3372549 , 0.17647059}
\definecolor{tblue}{HTML}{01847F}

\definecolor{retrain}{HTML}{D3D3D3}

\ifdefined\ShowNotes
  \newcommand{\colornote}[3]{{\color{#1}\bf{#2: #3}\normalfont}}
\else
  \newcommand{\colornote}[3]{}
\fi

\newcommand{\myparagraph}[1]{\vspace*{1.5pt}{\bf\noindent #1}}

\newcommand{\eat}[1]{} %

\newlength\savewidth

\makeatletter
\renewcommand{\paragraph}{%
  \@startsection{paragraph}{4}%
  {\z@}{0.3ex \@plus 1ex \@minus .1ex}{-1em}%
  {\normalfont\normalsize\bfseries}%
}
\makeatother

\newif\ifproofread

\usepackage{wrapfig}

\usepackage{amsmath,amssymb,amsthm}
\usepackage{placeins}  %
\usepackage{float}     %

\title{Moving Alphabet: A Controlled Study of Training Data for Text-to-Video Generation}

\author[1,2,*]{Amber Yijia Zheng}
\author[1]{Lu Liu}
\author[2]{Raymond A. Yeh}
\author[1]{Xi Yin}

\affiliation[1]{Meta Superintelligence Labs}
\affiliation[2]{Purdue University}

\contribution[*]{Work done during an internship at Meta Superintelligence Labs}

\definecolor{burntorange}{rgb}{0.81,.33,0}

\renewcommand{\cite}[1]{\citep{#1}}

\abstract{
Text-to-video generation has advanced significantly over the past five years through scaling of model size, data, and compute. Unlike model architecture, training data is often underexplored. Real-world data curation is complex and non-trivial, involving clip selection from raw videos and captioning to create video-text pairs for learning text-to-video mappings. We study how data distribution and caption quality impact text-to-video models. To enable controlled experiments, we introduce \textbf{Moving Alphabet}, a procedural testbed that renders letters with varying fonts, colors, sizes, and positions, moving in different directions and speeds against a black background. This design allows precise control over data distribution and caption quality by corrupting ground-truth metadata. Our experiments yield three findings: a) a diverse and balanced distribution of video content and duration is critical for generalization; b) caption quality significantly affects both model performance and training efficiency, suggesting that text-to-video models are bounded by video understanding capabilities; and c) classifier-free guidance and fine-tuning on high-quality data provide partial recovery from models trained on corrupted captions, but cannot fully compensate for poor pre-training data. We believe these insights can inform the development of large-scale text-to-video models, and we advocate for greater attention to the science of pre-training data.
}

\date{\today}
\correspondence{Amber Yijia Zheng at \email{amberrzheng@gmail.com}, Xi Yin at \email{yinxi@meta.com}}

\begin{document}

\maketitle
\section{Introduction}
\label{sec:intro}

Driven by architectural innovation, increased computation power, and the scaling of training datasets, Text-to-video (T2V) generation models have progressed from producing short artifact-laden clips~\cite{singer2022makeavideo, ho2022imagenvideo} to synthesizing complex scenes with synchronized audio~\cite{sora, polyak2024moviegen} that are often indistinguishable from real footage. 
Much of the existing literature has focused on model and architectural innovations, such as diffusion models~\cite{ho2020denoising, rombach2022high}, transformer architectures~\cite{vaswani2017attention, peebles2023scalable}, and scaling laws~\cite{kaplan2020scaling}, to explain these improvements. However, what is less discussed is the critical role of training data quality and composition. As the saying goes: ``garbage in, garbage out.'' In this paper, we study the effect of training data on the quality of video models, specifically focusing on how video data characteristics, \eg, content complexity and caption quality, impact the performance of T2V models. We also investigate whether techniques like classifier-free guidance (CFG) and fine-tuning can mitigate the effects of low-quality training data. These studies then offer guidelines on what to prioritize when curating training data for video models.

Briefly, a video data pipeline typically consists of two main stages: (a) {\bf sourcing and clipping} that selects videos and extracts short clips typically capped at 10-20s using model-driven heuristics; (b) {\bf video captioning} based on vision-language models to annotate each clip with detailed descriptions. Despite substantial progress in multimodal understanding models~\cite{yang2025qwen3,comanici2025gemini,team2025kimi}, for clip selection and captioning, it remains unclear how the video distribution and caption quality affect the model performance. Importantly, how much headroom remains in training data to further advance T2V generation? In this paper, we focus on studying the following aspects of the training videos: 
\begin{itemize}
    \item What is a suitable visual content complexity and duration distribution in the training data?
    \item How caption quality impacts model performance and training efficiency?
    \item Whether inference-time guidance (classifier-free guidance) or post-training fine-tuning on a small amount of high-quality data can recover performance for models pretrained on lower-quality data?
\end{itemize}

To systematically study these aspects, we need a testbed that allows precise control over video content complexity, duration, and caption quality. Working directly with real-world video data is impractical, so we introduce \textbf{Moving Alphabet}, a controlled video testbed where we render letters with varying fonts, colors, sizes, and positions, moving at different speeds and directions. This design enables precise control over the data distribution and allows us to generate captions with varying precision and recall by manipulating ground-truth metadata. We also propose novel, fine-grained evaluation metrics for assessing visual quality and prompt adherence.
{\bf Our experiments yield the following recommendations for curating T2V training data}: 
\begin{itemize}
    \item \textbf{Prioritize diverse and balanced data distributions.} Models trained on an equal mix of 1-letter, 2-letter, and 3-letter scenes match or exceed specialists trained on a single complexity with considerable margins, even when evaluated on the specialist's own data distribution. We observe the same trend for video duration mixing of 2-seconds, 4-seconds and 8-seconds. These findings recommend that practitioners aim for a diverse and balanced data distribution when collecting pre-training video data. We verify this for content complexity and clip duration, and expect the same principle to extend to other axes we did not study, such as resolution. %

    \item \textbf{Invest in high-quality captions from the start.} Model performance drops significantly and sharply with reductions in both precision (caption correctness) and recall (caption completeness), and precision is the more damaging axis: under a limited budget, correctness should be prioritized over completeness. Models trained on corrupted captions require up to $2$--$4\times$ more compute to match the validation loss of models trained on ground-truth captions. Although perfect precision might be achieved under low recall, full recall is fundamentally ill-defined for real videos because the desired level of detail varies. Given that T2V performance and controllability are primarily bounded by caption content and accuracy, we recommend prioritizing caption quality during data curation. More broadly, much of the recent progress in video generation can be traced to better captions from improved media-understanding models; while practitioners already recognize that captions matter, our results quantify how severely caption quality still bottlenecks performance, pointing to substantial remaining headroom in this direction. %

    \item \textbf{Do not rely on inference or post-training techniques to fix poor pre-training data.} While classifier-free guidance can improve prompt following to some degree regardless of caption quality, and fine-tuning on a small set of clean data can only recover up to 55\% of lost FG~PSNR for moderate corruption. %
    These results suggest that practitioners should invest in caption quality during the pre-training stage rather than attempting to recover performance post-hoc.
\end{itemize}

While our findings are derived from a controlled testbed, we believe they offer valuable insights that can guide real-world T2V data curation strategies. We hope this work brings attention to the yet understudied role of training data and encourages the research community to invest as much effort in data quality and composition as in architectural/model innovation.

\section{Related Work}
\label{sec:related}

\myparagraph{Video generation models.}
Diffusion-based video generation has advanced rapidly.
Early text-to-video systems such as Make-A-Video~\cite{singer2022makeavideo} and Imagen Video~\cite{ho2022imagenvideo} extended image diffusion to the temporal domain using cascaded or spatiotemporal architectures.
More recent models adopt latent diffusion with transformer backbones. \Eg, 
Latte~\cite{ma2024latte} uses a latent diffusion transformer,
CogVideoX~\cite{yang2024cogvideox} introduces an expert transformer design,
and Stable Video Diffusion~\cite{blattmann2023stablevideo} scales latent video models to large curated datasets.
Lumiere~\cite{bartal2024lumiere} proposes a space-time U-Net that generates entire video clips in a single pass.
At larger scale, Sora~\cite{sora}, Movie~Gen~\cite{polyak2024moviegen}, and HunyuanVideo~\cite{kong2024hunyuanvideo} demonstrate minute-long, high-definition generation, while Open-Sora~\cite{zheng2024opensora} provides an open reproduction.
These systems report strong qualitative results, but their training recipes, \ie, details of data composition, caption quality requirements, and mixing strategies, are described only at a high level. Our work differs by providing controlled experiments that isolate the effect of each training data axis.

\myparagraph{Training data curation for generative models.}
Data quality has been shown to matter as much as scale for image generation.
DALL$\cdot$E~3 by~\citet{betker2023dalle3} demonstrated that recaptioning training images with a dedicated captioner dramatically improves text-image alignment; often more than scaling the dataset.
On the dataset side, LAION-5B~\cite{schuhmann2022laion5b} introduced large-scale filtering pipelines, and DataComp~\cite{datacomp} systematically benchmarked data selection strategies for CLIP training.
For video, Panda-70M~\cite{chen2024panda70m} and InternVid~\cite{wang2024internvid} build large captioned video datasets using multiple cross-modality teachers, while ShareGPT4Video~\cite{chen2025sharegpt4video} and LLaVA-NeXT~\cite{liu2024llavanext} improve video understanding through better captions.
Stable Video Diffusion~\cite{blattmann2023stablevideo} provides the most comprehensive data curation study for video generation to date, showing through systematic ablations that careful filtering and a multi-stage curation strategy are decisive for downstream quality.
VideoCrafter2~\cite{chen2024videocrafter2} explicitly studies how to overcome data limitations for high-quality video diffusion.
Note, these efforts focus on scaling and filtering real-world data, where content, caption quality, and visual fidelity are entangled. In this work, we leverage a controlled testbed that allows us to vary each axis independently and measure its importance. 

\myparagraph{Synthetic testbeds for understanding generative models.}
Controlled datasets with known ground truth have been instrumental for understanding representation learning.
CLEVR~\cite{johnson2017clevr} tests compositional visual reasoning, dSprites~\cite{matthey2017dsprites} and 3D~Shapes~\cite{kim2018shapes3d} provide disentanglement benchmarks with independently varying factors, and T2I-CompBench~\cite{huang2023t2icompbench} evaluates compositional text-to-image generation.
In the language domain, TinyStories~\cite{eldan2023tinystories} shows that small models trained on synthetic stories can produce coherent language, and Physics of Language Models~\cite{allenzhu2024physics} uses synthetic data to study knowledge storage and extraction.
Muennighoff et al.~\cite{muennighoff2024scaling} study data scaling laws for language models under controlled data constraints.
Our Moving Alphabet testbed extends this methodology to video generation: every attribute is known, captions can be corrupted along precise axes, and evaluation is exact rather than perceptual.

\myparagraph{Curriculum and data mixing.}
Curriculum learning~\cite{bengio2009curriculum} proposes training on easy examples first, while recent work on data mixing for LLMs studies how the composition of training data affects final performance.
DoReMi~\cite{xie2023doremi} learns domain weights to speed up pretraining, The Pile~\cite{gao2020pile} combines diverse text sources into a single training corpus, and Data Mixing Laws~\cite{ye2024datamixing} formalize how to predict performance from mixture proportions.
Scaling laws~\cite{kaplan2020scaling} characterize how model and data size trade off, but say little about data \emph{composition} at fixed scale.
In video generation, progressive training from low to high resolution is common practice~\cite{sora,polyak2024moviegen}, but the effect of mixing different scene complexities or video lengths has not been systematically studied.
Our complexity and length mixing experiments (\S\ref{sec:complexity-mixing}, \S\ref{sec:temporal-mixing}) provide the first controlled evidence that equal mixing across content axes consistently outperforms specialist or biased training.

\begin{figure*}[t]
    \centering
    \begin{subfigure}{\textwidth}
        \centering
        \includegraphics[width=0.158\textwidth]{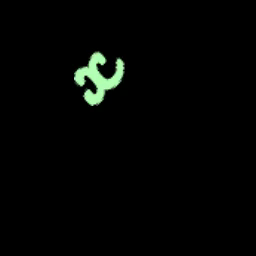}\hfill
        \includegraphics[width=0.158\textwidth]{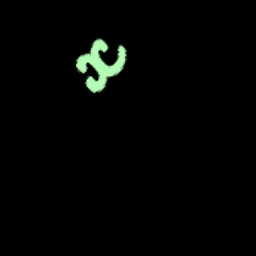}\hfill
        \includegraphics[width=0.158\textwidth]{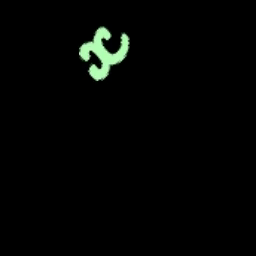}\hfill
        \includegraphics[width=0.158\textwidth]{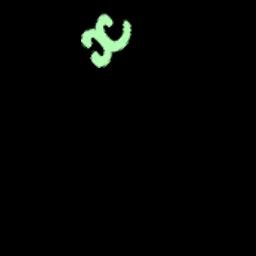}\hfill
        \includegraphics[width=0.158\textwidth]{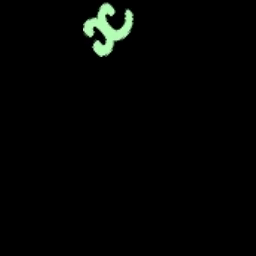}\hfill
        \includegraphics[width=0.158\textwidth]{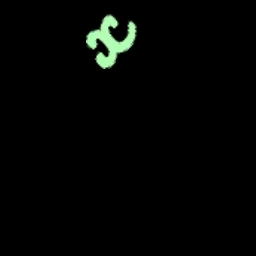}
        \caption{\textbf{1-letter.} ``A medium light green lowercase x in handwritten font, rotated 315 degrees. It starts at (0.29, 0.19) and moves at 291.4 degrees at speed 4.3, ending at (0.34, 0.07). It bounces off the top wall.''}
        \label{fig:ex-1l}
    \end{subfigure}

    \vspace{3pt}
    \begin{subfigure}{\textwidth}
        \centering
        \includegraphics[width=0.158\textwidth]{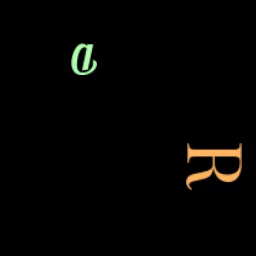}\hfill
        \includegraphics[width=0.158\textwidth]{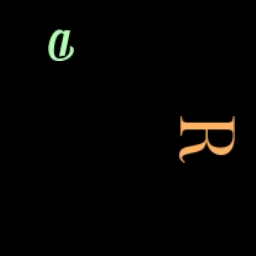}\hfill
        \includegraphics[width=0.158\textwidth]{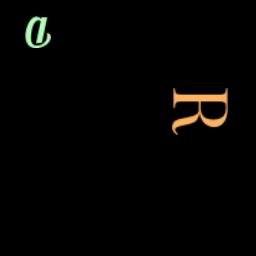}\hfill
        \includegraphics[width=0.158\textwidth]{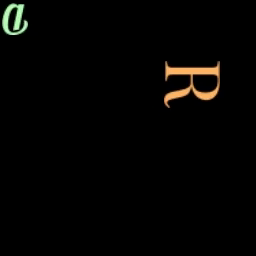}\hfill
        \includegraphics[width=0.158\textwidth]{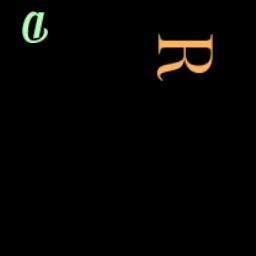}\hfill
        \includegraphics[width=0.158\textwidth]{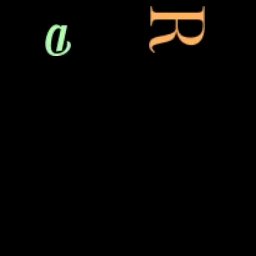}
        \caption{\textbf{2-letter.} ``A scene with 2 letters. Letter~1: A medium orange uppercase R in serif font, rotated 90 degrees. It starts at (0.73, 0.56) and moves at 254.3 degrees at speed 9.5, ending at (0.57, 0.01). It bounces off the top wall. Letter~2: A small light green uppercase D in decorative font, rotated 180 degrees. It starts at (0.28, 0.17) and moves at 216.8 degrees at speed 8.9, ending at (0.21, 0.11). It bounces off the left and top walls.''}
        \label{fig:ex-2l}
    \end{subfigure}

    \vspace{3pt}
    \begin{subfigure}{\textwidth}
        \centering
        \includegraphics[width=0.158\textwidth]{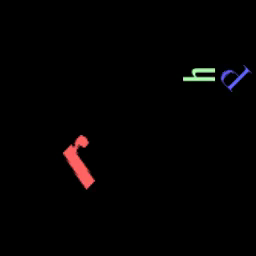}\hfill
        \includegraphics[width=0.158\textwidth]{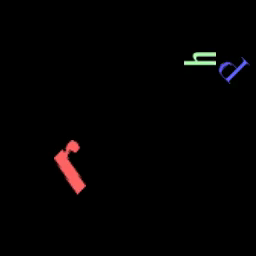}\hfill
        \includegraphics[width=0.158\textwidth]{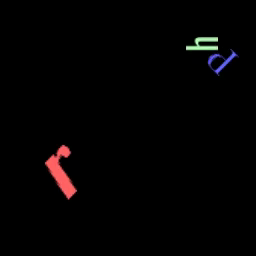}\hfill
        \includegraphics[width=0.158\textwidth]{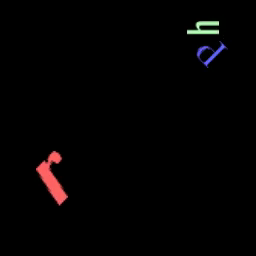}\hfill
        \includegraphics[width=0.158\textwidth]{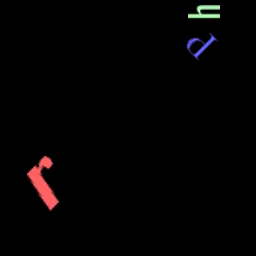}\hfill
        \includegraphics[width=0.158\textwidth]{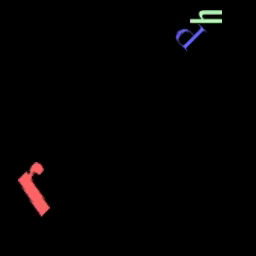}
        \caption{\textbf{3-letter.} ``A scene with 3 letters. Letter~1: A medium red lowercase r in decorative font, rotated 315 degrees. It starts at (0.25, 0.53) and moves at 149.0 degrees at speed 3.5, ending at (0.06, 0.64). Letter~2: A small light green lowercase h in condensed font, rotated 270 degrees. It starts at (0.71, 0.27) and moves at 278.4 degrees at speed 5.2, ending at (0.74, 0.06). It bounces off the top wall. Letter~3: A small blue uppercase P in serif font, rotated 225 degrees. It starts at (0.86, 0.25) and moves at 222.0 degrees at speed 4.5, ending at (0.67, 0.08). It bounces off the right wall.''}
        \label{fig:ex-3l}
    \end{subfigure}

    \caption{%
        Example videos from the Moving Alphabet dataset at three complexity levels, one per row. Frames are shown left to right at $t{=}0,3,6,9,12,15$ of the 17-frame clips ($256{\times}256$, 8\,fps). Each row's ground-truth caption, which describes every letter's appearance and motion, is shown beneath it.
    }
    \label{fig:dataset_examples}
\end{figure*}

\section{Moving Alphabet: A Controlled Video Generation Testbed}
\label{sec:testbed}

Studying how training data influences T2V models requires precise control over video content and caption quality. These conditions are difficult to study with real-world data, where ground truth is ambiguous and confounding factors are unavoidable. We therefore introduce \textbf{Moving Alphabet}, a procedural testbed that renders letters moving against a near-black background. Because all attributes are sampled from known distributions, we can {\bf (a)} introduce precise, measurable corruption into captions, {\bf (b)} compose any target data distribution, and {\bf (c)} evaluate models against exact per-video ground truth.
\figref{fig:dataset_examples} shows example clips together with their ground-truth captions at each complexity level.

\subsection{Data Generation}
\label{sec:data-gen}
We construct the Moving Alphabet dataset by placing one to three letters with varying fonts, colors, sizes, and rotations on a black background, and animating them to move in different directions with different speeds. The main benefits of this procedural approach are that datasets can be scaled to arbitrary size and composed into controlled mixtures, enabling controlled experiment setups. Specifically, we systematically vary video content complexity, video duration, and caption quality; details below.

\myparagraph{Content complexity.}
Each video contains one to three letters, which we refer to as $1$-letter ($1$L), $2$-letter ($2$L), and $3$-letter ($3$L) scenes. Each letter is defined by seven independently sampled attributes covering both appearance and motion:
\begin{itemize}
    \item \textbf{Appearance}: letter identity is sampled from $52$ classes consisting of $26$ uppercase and $26$ lowercase letters; font is sampled from $8$ typefaces, color from $10$ options, size from $3$ levels (small, medium, and large), and rotation from $8$ angles. Full attribute specifications are in Appendix~\ref{app:attributes}.
    \item \textbf{Motion}: each letter is initialized at a random position with a speed sampled uniformly in $[3,10]$ pixels per frame and a heading sampled uniformly in $[0^\circ,360^\circ)$. The letter moves at constant speed and reflects off the canvas boundaries, with every bounce recorded in the metadata. We summarize the motion by a single direction attribute, the angle of the net start-to-end displacement; since it is measured from the realized trajectory, it already accounts for any bounces rather than reflecting the initial heading.
\end{itemize}
We sample each letter independently. Increasing the number of letters introduces overlapping trajectories and shared canvas space; where letters overlap, they are alpha-composited~\cite{porter1984compositing} in a fixed index order, so a higher-indexed letter occludes a lower-indexed one.

\myparagraph{Temporal complexity.}
Video length is independently configurable to roughly $2$, $4$, or $8$ seconds, corresponding to $17$, $33$, or $65$ frames\footnote{The causal temporal VAE compresses frames $4\times$ in time and encodes the first frame independently, so it expects $4n{+}1$ input frames; we therefore use $17$, $33$, and $65$ frames rather than $16$, $32$, and $64$.} at $8$ fps and $256\times256$ resolution. Longer videos typically contain more boundary bounces and exhibit more complex trajectory patterns.

\myparagraph{Captions.} Each video is paired with a templated natural-language caption generated from its metadata. The captions describe all appearance attributes, start and end positions using normalized coordinates, movement direction as an exact angle, speed, and any boundary bounces. Figure~\ref{fig:dataset_examples} shows some examples.

\subsection{Model Architecture and Training}
\label{sec:architecture}

\myparagraph{Architecture.}
We adopt a standard MMDiT architecture~\cite{mmdit} trained with flow matching~\cite{lipman2023flow}. The model contains $800$M parameters, consisting of $36$ transformer blocks, with a hidden dimension of $1024$ and $16$ attention heads. For text conditioning, we use two frozen text encoders: T5-XXL~\cite{raffel2020t5} with $4096$-dimensional embeddings and CLIP ViT-L/14~\cite{radford2021clip} with $768$-dimensional embeddings. Their token sequences are concatenated along the sequence dimension and jointly attended to by the transformer.
Video latents are obtained using a pretrained causal $3$D VAE that encodes the first frame independently and applies $8\times$ spatial and $4\times$ temporal compression, followed by spatial patchification with a $2\times2$ patch size and temporal patch size as $1$. We adopt this standard recipe~\cite{kong2024hunyuanvideo,yang2024cogvideox,polyak2024moviegen} so our findings transfer to deployed systems, at a moderate $800$M scale that keeps hundreds of controlled experiments affordable.

\myparagraph{Baseline.}
All models are trained from scratch on $8\times$A100 GPUs. The base configuration uses $150$K videos with $17$ frames at $8$ fps and $256\times256$ resolution, converging in approximately two days. All metrics plateau by epoch $100$ (see \figref{fig:convergence} in the appendix); we use this checkpoint for evaluation and as the starting point for finetuning experiments unless otherwise specified. During training, we randomly drop the caption with a probability $0.1$ to enable classifier-free guidance at inference. Full training hyperparameters are in Appendix~\ref{app:training-details}.

\subsection{Evaluation}
\label{sec:evaluation}

A held-out set of $6{,}000$ videos ($2{,}000$ per complexity level), each paired with its ground-truth caption, is used for evaluation.
Unless otherwise specified, we report results at CFG\,=\,1.0 to exclude the effect of guidance scale; \S\ref{sec:cfg_compensation} studies how CFG interacts with caption quality.

Standard video metrics such as Fréchet Video Distance (FVD)~\cite{unterthiner2019fvd} rely on I3D features pretrained on natural videos, which transfer poorly to our synthetic domain. Since every video instead comes with exact ground-truth metadata, we use two complementary metric families that exploit it:
\begin{itemize}
    \item \textbf{Pixel accuracy}: foreground PSNR (FG~PSNR) and foreground MSE (FG~MSE), computed on the letter region only. These capture both visual fidelity and semantic correctness, as incorrect colors or positions increase pixel errors relative to the GT rendering. We report FG~MSE scaled by $10^{3}$ for readability.
    \item \textbf{Attribute accuracy}: we classify each attribute ($A_\text{color}$, $A_\text{size}$, $A_\text{dir}$, $\rho_\text{speed}$) from generated video pixels and compare against ground-truth metadata. For multi-letter videos, letters are localized using ground-truth trajectory bounding boxes to avoid cross-letter interference. Full metric definitions are in Appendix~\ref{app:eval-details}.
\end{itemize}

\begin{table}[t]
\centering
\caption{%
\textbf{(a)} Metric calibration: attribute accuracy evaluated on raw ground-truth videos, confirming that our classifiers are near-perfect on uncompressed data.
\textbf{(b)} VAE reconstruction ceiling: all metrics evaluated after encode$\to$decode round-trip, establishing the upper bound for generated video evaluation.
}
\label{tab:vae_ceiling}

\vspace{0.3em}
\small
\setlength{\tabcolsep}{4pt}

\begin{minipage}[t]{0.45\linewidth}
\centering
\textsc{(a) Metric Calibration (Raw)}
\vspace{0.2em}

\begin{tabular}{@{}l ccc@{}}
\toprule
& 1L & 2L & 3L \\
\midrule
Color\,$\uparrow$    & 0.995 & 0.975 & 0.956 \\
Dir\,$\uparrow$      & 0.996 & 0.992 & 0.990 \\
Size\,$\uparrow$     & 1.000 & 1.000 & 1.000 \\
Speed\,$\uparrow$    & 0.996 & 0.995 & 0.994 \\
\bottomrule
\end{tabular}
\end{minipage}%
\hfill
\begin{minipage}[t]{0.52\linewidth}
\centering
\textsc{(b) VAE Reconstruction Ceiling}
\vspace{0.2em}

\begin{tabular}{@{}l ccc@{}}
\toprule
& 1L & 2L & 3L \\
\midrule
Color\,$\uparrow$    & 0.987 & 0.967 & 0.946 \\
Dir\,$\uparrow$      & 0.996 & 0.993 & 0.990 \\
Size\,$\uparrow$     & 1.000 & 1.000 & 1.000 \\
Speed\,$\uparrow$    & 0.998 & 0.996 & 0.995 \\
\midrule
FG MSE ($\times 10^{3}$)\,$\downarrow$ & 5.6 & 4.7 & 4.6 \\
FG PSNR (dB)\,$\uparrow$ & 22.5 & 23.2 & 23.4 \\
PSNR (dB)\,$\uparrow$    & 41.1 & 38.1 & 36.2 \\
\bottomrule
\end{tabular}
\end{minipage}

\vspace{-0.5em}
\end{table}

Our per-attribute scores are produced by automated classifiers, which we apply to generated videos to measure what the model actually produced. A generated video carries no attribute label of its own, which must be measured from the pixels, and an unreliable classifier would make a low score reflect the metric rather than the model. We therefore calibrate the classifiers on ground-truth videos with known attributes, thereby demonstrating the classifiers' correctness. As shown in~\tabref{tab:vae_ceiling}(a), the classifiers recover the true attributes near-perfectly, the lowest being color at 0.956 on 3L scenes. Hence, the classifiers are reliable instruments, and any error they later report on generated videos reflects the model, not measurement noise. The residual drop comes from the inter-letter occlusion described in \S\ref{sec:data-gen}: where letters overlap, the occluding letter's pixels enter the occluded one's ROI and bias the median color. The effect grows with the number of letters, with the color ceiling decreasing from $0.995$ (1L) to $0.975$ (2L) and $0.956$ (3L).

Next, we also consider the effect of VAE on the metrics. Every generated video is decoded by the same frozen, lossy VAE, which caps the pixel quality any model can reach, so FG~PSNR is only meaningful relative to that cap. \tabref{tab:vae_ceiling}(b) measures every metric on the VAE encode--decode of the ground-truth videos, establishing this ceiling for FG~PSNR of $22$--$23$\,dB and for FG~MSE of $4.6$--$5.6\times 10^{-3}$. %
Our best models reach FG~PSNR of $13$--$14$\,dB, $9$--$10$\,dB below this ceiling; hence, our experiments are therefore not bottlenecked by VAE quality, and the differences we observe across settings reflect the quality of the diffusion model itself.

\begin{table*}[t]
\centering
\caption{%
Effect of training data complexity mixing on attribute accuracy and pixel quality.
Models are trained on different distributions over 1L/2L/3L scenes with same total data budget and evaluated at all complexity levels with CFG\,=\,1.0.
\textbf{Bold}: best per column; \underline{underline}: second best.
}
\label{tab:complexity_mixing}

\vspace{0.3em}
\small
\setlength{\tabcolsep}{3pt}

\textsc{(a) Attribute Accuracy}
\vspace{0.2em}

\resizebox{\textwidth}{!}{%
\begin{tabular}{@{}l cccc cccc cccc cccc@{}}
\toprule
& \multicolumn{4}{c}{\textit{Eval: 1-Letter}}
& \multicolumn{4}{c}{\textit{Eval: 2-Letter}}
& \multicolumn{4}{c}{\textit{Eval: 3-Letter}}
& \multicolumn{4}{c}{\textit{Avg}} \\
\cmidrule(lr){2-5} \cmidrule(lr){6-9} \cmidrule(lr){10-13} \cmidrule(lr){14-17}
Training Data
& Col\,$\uparrow$ & Dir\,$\uparrow$ & Size\,$\uparrow$ & Spd\,$\uparrow$
& Col\,$\uparrow$ & Dir\,$\uparrow$ & Size\,$\uparrow$ & Spd\,$\uparrow$
& Col\,$\uparrow$ & Dir\,$\uparrow$ & Size\,$\uparrow$ & Spd\,$\uparrow$
& Col\,$\uparrow$ & Dir\,$\uparrow$ & Size\,$\uparrow$ & Spd\,$\uparrow$ \\
\midrule
Pure 1L
& \underline{.979} & \underline{.983} & \textbf{1.00} & \textbf{.998}
& .309 & .467 & .283 & .450
& .165 & .357 & .172 & .282
& .484 & .602 & .485 & .577 \\
Pure 2L
& .610 & .975 & .941 & .915
& .878 & \underline{.984} & .972 & \underline{.994}
& .532 & .761 & .635 & .504
& .673 & .906 & .849 & .804 \\
Pure 3L
& .379 & .485 & .346 & .281
& .894 & .982 & .995 & .992
& .886 & .980 & .995 & \underline{.993}
& .720 & .816 & .779 & .755 \\
\midrule
Mix Equal
& \textbf{.981} & \textbf{.987} & \textbf{1.00} & \underline{.997}
& \textbf{.950} & \textbf{.987} & \textbf{1.00} & \textbf{.997}
& \underline{.918} & \underline{.981} & \textbf{1.00} & \textbf{.996}
& \textbf{.950} & \textbf{.985} & \textbf{1.00} & \textbf{.997} \\
Heavy-1L
& .970 & .943 & \underline{.999} & .950
& \textbf{.950} & .977 & \underline{.998} & \underline{.994}
& \underline{.918} & .975 & \underline{.999} & \underline{.993}
& \underline{.946} & .965 & \underline{.998} & \underline{.979} \\
Heavy-3L
& .939 & .949 & .919 & .915
& \underline{.946} & \underline{.984} & \underline{.998} & \textbf{.997}
& \textbf{.920} & \textbf{.982} & .995 & \textbf{.996}
& .935 & \underline{.972} & .970 & .969 \\
\bottomrule
\end{tabular}%
}

\vspace{1em}

\textsc{(b) Pixel Quality}
\vspace{0.2em}

\begin{tabular}{@{}l cc cc cc cc@{}}
\toprule
& \multicolumn{2}{c}{\textit{Eval: 1-Letter}}
& \multicolumn{2}{c}{\textit{Eval: 2-Letter}}
& \multicolumn{2}{c}{\textit{Eval: 3-Letter}}
& \multicolumn{2}{c}{\textit{Avg}} \\
\cmidrule(lr){2-3} \cmidrule(lr){4-5} \cmidrule(lr){6-7} \cmidrule(lr){8-9}
Training Data
& \small{FG MSE${\downarrow}$} & \small{FG PSNR${\uparrow}$}
& \small{FG MSE${\downarrow}$} & \small{FG PSNR${\uparrow}$}
& \small{FG MSE${\downarrow}$} & \small{FG PSNR${\uparrow}$}
& \small{FG MSE${\downarrow}$} & \small{FG PSNR${\uparrow}$} \\
& \multicolumn{8}{c}{\scriptsize FG MSE values are $\times 10^{3}$; FG PSNR in dB.} \\
\midrule
Pure 1L
& 70.1 & 12.4
& 383.4 & 4.3
& 402.0 & 4.1
& 285.2 & 6.9 \\
Pure 2L
& 168.5 & 8.6
& 85.8 & 11.7
& 265.3 & 6.1
& 173.2 & 8.8 \\
Pure 3L
& 187.7 & 8.2
& 109.7 & 10.4
& 113.7 & 10.1
& 137.0 & 9.6 \\
\midrule
Mix Equal
& \textbf{63.9} & \textbf{12.8}
& \textbf{55.4} & \textbf{13.1}
& \textbf{56.0} & \textbf{12.9}
& \textbf{58.4} & \textbf{12.9} \\
Heavy-1L
& 136.0 & 9.6
& 104.9 & 10.4
& 90.1 & 10.8
& 110.3 & 10.3 \\
Heavy-3L
& \underline{68.9} & \underline{12.5}
& \underline{62.2} & \underline{12.6}
& \underline{59.8} & \underline{12.7}
& \underline{63.6} & \underline{12.6} \\
\bottomrule
\end{tabular}

\vspace{-0.5em}
\end{table*}

\FloatBarrier
\section{Investigation on Data Distribution}
\label{sec:content-complexity}

We first investigate the data collection choices on how we should balance different types of training data. We study two axes: content complexity (simple vs. complex scenes) and video length (short vs. long clips). Throughout this section, we hold caption quality fixed at ground truth, so that the only variable across conditions is the data composition under study.

\FloatBarrier
\subsection{Content Complexity}
\label{sec:complexity-mixing}
\myparagraph{Experimental setup.} The choice of what data to collect is naturally tied to the target evaluation scenarios. However, even when the evaluation distribution is known, it remains unclear whether the training data should match that distribution or whether another data distribution may yield better performance. We study this question by constructing training sets with different mixtures of 1L, 2L, and 3L videos and evaluate on different evaluation sets while keeping the total training set size fixed.

Following the baseline setting of $150$K $2$-second videos with $17$ frames at $8$ FPS, and ground-truth captions, we construct $6$ different training sets.
Three single-complexity conditions using pure 1L, 2L, or 3L videos, and three mixtures. The equal mix contains $50$K videos at each level. The Heavy-1L mix has $100$K $1$L with $25$K each of $2$L and $3$L, while the Heavy-3L mix has $100$K 3L and $25$K each of 1L and 2L.
We evaluate each model on held-out videos at all three complexity levels.
The results are shown in~\tabref{tab:complexity_mixing} using our proposed evaluation metrics. We summarized our findings below.

\myparagraph{Out-of-distribution collapse.} As expected, models trained on a single complexity level fail when evaluated on unseen complexities. Pure 1L training achieves only $0.357$ direction accuracy and $4.1$\,dB FG~PSNR on 3L scenes, while Pure 3L drops to $0.485$ direction accuracy and $8.2$\,dB FG~PSNR on 1L scenes. The degradation is not graceful, which suggests that the compositional skills do not emerge from single-object training, and per-object accuracy is lost when training exclusively on complex scenes where multiple overlapping objects dilute the supervision signal for each object.

\myparagraph{Equal mixing matches or exceeds specialists.} Empirically, we found that Mix Equal not only generalizes across complexity levels but also matches or exceeds the specialist models when evaluated on their own distributions. On $3$L evaluation, Mix Equal reaches $12.9$\,dB FG~PSNR versus Pure 3L's $10.1$\,dB, a gap of nearly $2.8$\,dB (\tabref{tab:complexity_mixing}). We hypothesize that the cause of this is compositional transfer. That is, simple scenes provide clean, unambiguous supervision for per-object primitives like color, direction, and size, while complex scenes introduce spatial reasoning between objects. These primitives then compose into better multi-object generation.
This suggests that maintaining a diverse and balanced datamix is important in real-world text-to-video generation.

\myparagraph{Balanced mixing outperforms biased mixing.} Heavy-1L underperforms Mix Equal even on $1$L evaluation despite containing more $1$L data, with a 3.2\,dB FG~PSNR gap (\tabref{tab:complexity_mixing}). This suggests that oversampling a single complexity level does not translate to better performance, even on that level's own distribution. Heavy-3L is competitive with Mix Equal on 3L metrics but sacrifices 1L and 2L quality.

\myparagraph{\color{burntorange} Takeaways.}
Training on a single scene complexity fails to generalize. Models collapse out-of-distribution, and compositional skills do not emerge from single-object training alone. Oversampling a single complexity level does not improve performance even on that level's own evaluation set.
For practitioners, we recommend maintaining a balanced mix of simple and complex scenes rather than filtering toward either extreme. Intuitively, simple scenes provide clean per-object supervision, complex scenes build spatial reasoning, and together they produce stronger multi-object generation at no extra data cost.

\FloatBarrier
\subsection{Video Length}
\label{sec:temporal-mixing}
\begin{table*}[t]
\centering
\caption{%
Effect of training video length distribution on pixel quality.
All conditions use 150K videos but differ in total tokens due to varying durations.
Metrics are averaged over 1L/2L/3L complexity levels (CFG\,=\,1.0).
\textbf{Bold}: best per column; \underline{underline}: second best. 
}
\label{tab:length_mixing}

\vspace{0.3em}
\small
\setlength{\tabcolsep}{3.6pt}

\resizebox{\textwidth}{!}{%
\begin{tabular}{@{}l r cc cc cc cc@{}}
\toprule
& & \multicolumn{2}{c}{\textit{Eval: 2s (17f)}}
& \multicolumn{2}{c}{\textit{Eval: 4s (33f)}}
& \multicolumn{2}{c}{\textit{Eval: 8s (65f)}}
& \multicolumn{2}{c}{\textit{Avg}} \\
\cmidrule(lr){3-4} \cmidrule(lr){5-6} \cmidrule(lr){7-8} \cmidrule(lr){9-10}
Training Data & Tok.\,(B)
& \small{FG MSE${\downarrow}$} & \small{FG PSNR${\uparrow}$}
& \small{FG MSE${\downarrow}$} & \small{FG PSNR${\uparrow}$}
& \small{FG MSE${\downarrow}$} & \small{FG PSNR${\uparrow}$}
& \small{FG MSE${\downarrow}$} & \small{FG PSNR${\uparrow}$} \\
& & \multicolumn{8}{c}{\scriptsize FG MSE values are $\times 10^{3}$; FG PSNR in dB.} \\
\midrule
Pure 2s & 19.2
& \textbf{58.4} & \textbf{12.9}
& 354.9 & 4.7
& 370.1 & 4.5
& 261.1 & 7.4 \\
Pure 4s & 34.6
& 312.4 & 5.3
& 101.3 & 11.1
& 355.8 & 4.7
& 256.5 & 7.0 \\
Pure 8s & 65.3
& 297.6 & 5.5
& 282.8 & 5.8
& 101.6 & 10.9
& 227.3 & 7.4 \\
\midrule
Mix Equal & 39.7
& \underline{65.6} & \underline{12.4}
& \textbf{67.1} & \textbf{12.3}
& \underline{71.7} & \underline{12.0}
& \textbf{68.1} & \textbf{12.2} \\
Heavy-2s & 29.4
& 74.0 & 12.0
& \underline{72.4} & \underline{12.0}
& 79.7 & 11.6
& \underline{75.4} & \underline{11.9} \\
Heavy-8s & 52.5
& 223.8 & 6.6
& 81.3 & 11.2
& \textbf{64.6} & \textbf{12.5}
& 123.2 & 10.1 \\
\bottomrule
\end{tabular}%
}

\vspace{-0.5em}
\end{table*}

\myparagraph{Experimental setup.}
Video datasets in the wild span a wide range of durations. %
Longer videos contain richer temporal dynamics but are more expensive to process; shorter clips may lack complex motion patterns.
We study how the mixture of training video lengths affects generation quality across all durations.

We train six variants with $150$K videos each that vary the mixture of $2$-second, $4$-second, and $8$-second clips. All conditions share the same data budget of $150$K videos and $100$ epochs, matching the number of training clips each model sees rather than the token count or compute, which necessarily differ across clip lengths.
Three models are trained on a single duration each. Pure $2$s is trained on $19.2$B tokens, whereas Pure $8$s is trained on $65.3$B tokens, a $3.4\times$ difference.
The three mixed models combine all three durations. Mix Equal uses an equal number of examples per duration, with $50$K examples each, for a total of $39.7$B tokens.
Heavy-$2$s allocates $100$K examples to $2$-second videos and $25$K examples each to $4$- and $8$-second videos, totaling $29.4$B tokens. Heavy-$8$s uses the reverse allocation, totaling $52.5$B tokens.
All models are evaluated on all three video lengths. We report pixel quality (FG~MSE and FG~PSNR) rather than per-attribute accuracy, since direction is defined from net displacement and degenerates for longer clips where bounces return the letter near its origin\footnote{This reflects a limitation of summarizing motion by a single per-sample direction; richer per-segment motion metrics could further improve the evaluation for longer or more complex scenes, though the effect is minor for $2$-second clips.}; pixel metrics stay length-comparable and already reflect attribute errors.
\tabref{tab:length_mixing} reports pixel quality metrics averaged over complexity levels at CFG\,=\,1.0. %

\myparagraph{Length-specific training does not generalize.}
As expected, models trained on a single duration perform well only at their training duration.
For example, pure $4$s achieves $11.1$\,dB FG~PSNR on $4$-second videos but drops below $5.3$\,dB on other durations (\tabref{tab:length_mixing}).
The model memorizes the training video duration distribution rather than learning temporal representations that transfer across durations.

\myparagraph{Equal mixing provides balanced performance.}
Mix Equal achieves 12.0--12.4\,dB FG~PSNR across all three evaluation lengths, staying within 0.5\,dB of the best pure model at each duration.
It trades 0.5\,dB at 2s, where Pure 2s achieves 12.9\,dB, for gains of 1.2\,dB at 4s and 1.1\,dB at 8s over the specialists (\tabref{tab:length_mixing}.
Averaged across all three lengths, Mix Equal achieves 12.2\,dB, clearly the best overall (next best is Heavy-2s at 11.9\,dB).
This is not simply a compute effect: Mix Equal uses $39.7$B tokens, comparable to Pure $4$s at $34.6$B and far less than Pure 8s at 65.3B, yet outperforms both at 4s and 8s.
Exposure to diverse temporal dynamics enables generalization across durations.
Relative to content complexity, video length shows weaker cross-condition transfer: single-duration specialists lose about $6.5$\,dB out of domain, versus about $4.5$\,dB for single-complexity specialists, making a balanced mix even more important here.

\myparagraph{Biased mixing excels on majority durations but drops a minority one.}
Heavy-2s degrades gracefully, maintaining FG~PSNR above 11.6\,dB at all lengths, including the minority 8-second duration.
Heavy-8s achieves the best 8-second quality (12.5\,dB), surpassing even the 8-second specialist Pure 8s (10.9\,dB) by 1.6\,dB.
However, it catastrophically fails on 2-second videos (6.6\,dB), the least-represented duration in its training mix.
This instability suggests that heavy oversampling of long videos creates fragile optimization dynamics where the model cannot reliably learn all shorter durations.
By contrast, Mix Equal and Heavy-2s are stable across all lengths.

\myparagraph{\color{burntorange} Takeaways.}
Training on a single video length fails to generalize: models memorize the training duration rather than learning transferable temporal representations.
For practitioners curating real-world video datasets, we recommend avoiding duration-biased filtering and instead sampling roughly equal numbers of short, medium, and long clips; this yields broad generalization at lower token cost than training on long videos alone, and avoids the sharp short-clip degradation seen with heavy long-video oversampling.

\FloatBarrier
\section{Investigation on  Caption Quality}
\label{sec:caption-quality}

Text-to-video models learn the mapping from text to video through the training captions.
In practice, captions from VLMs or alt-text vary widely in accuracy and completeness, yet the effect of the caption quality on generation performance is poorly understood. 
Given the nature of our Moving Alphabet dataset, we can easily control the precision and recall of the captions to study their impact on video generation performance.

\FloatBarrier
\subsection{Precision vs.\ Recall: Which Matters More?}
\label{sec:precision-vs-recall}
\begin{figure*}[t]
    \centering
    \includegraphics[width=0.55\textwidth]{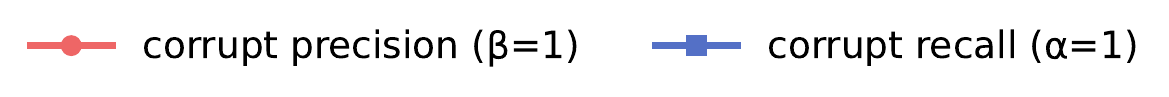}\\[2pt]
    \begin{subfigure}[t]{0.32\textwidth}
        \centering
        \includegraphics[width=\linewidth]{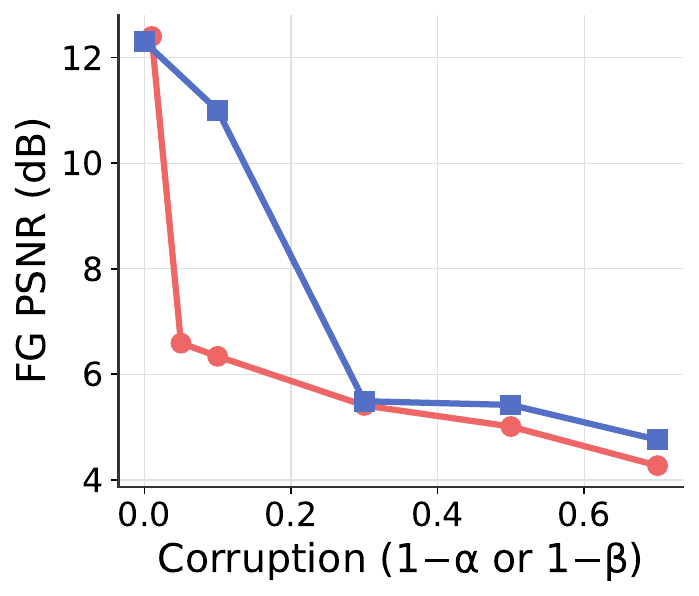}
        \caption{FG~PSNR}
        \label{fig:fg_psnr}
    \end{subfigure}\hfill
    \begin{subfigure}[t]{0.32\textwidth}
        \centering
        \includegraphics[width=\linewidth]{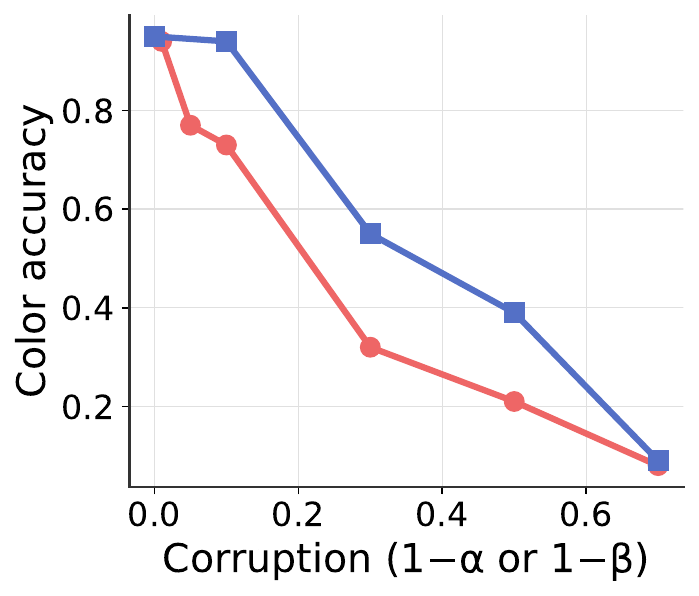}
        \caption{Color accuracy}
        \label{fig:color}
    \end{subfigure}\hfill
    \begin{subfigure}[t]{0.32\textwidth}
        \centering
        \includegraphics[width=\linewidth]{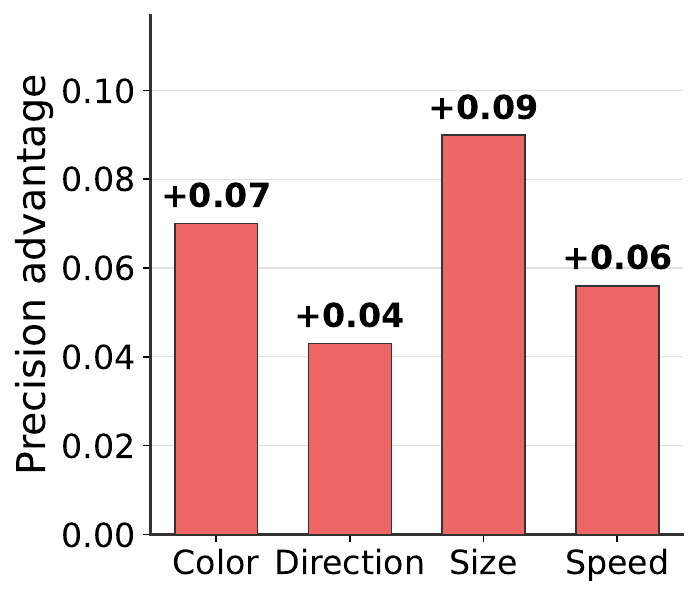}
        \caption{Precision advantage}
        \label{fig:precision_advantage}
    \end{subfigure}
    \caption{%
        Effect of caption precision ($\alpha$) and recall ($\beta$) on video generation quality at CFG\,=\,1.0.
        \textbf{(a, b)} Each curve corrupts one axis while holding the other perfect: the red curve degrades precision ($\beta{=}1$, decreasing $\alpha$) and the blue curve degrades recall ($\alpha{=}1$, decreasing $\beta$). For both FG~PSNR and color accuracy, corrupting precision causes the sharper drop, so precision is the more important axis; color collapses to near chance below $\alpha \leq 0.5$.
        \textbf{(c)} Precision advantage per attribute, an iso-budget swap. Let $M(\alpha,\beta)$ be an attribute's accuracy. For every pair of corruption levels $x{>}y$ shared by both axes, we compare giving the higher level to precision, $M(\alpha{=}x,\beta{=}y)$, against giving it to recall, $M(\alpha{=}y,\beta{=}x)$; the bar is the mean of $M(\alpha{=}x,\beta{=}y){-}M(\alpha{=}y,\beta{=}x)$ over all such pairs. It is positive for all four attributes, so under a fixed caption-quality budget, spending it on precision rather than recall improves every attribute, confirming that precision is the primary driver of attribute controllability.
        Full $\alpha \times \beta$ heatmaps and the per-pair swap scatter are in the appendix (\figref{fig:caption_additional}, \figref{fig:swap_scatter}).
    }
    \label{fig:precision_recall}
\end{figure*}

\myparagraph{Experimental setup.}
\label{sec:caption-model}
Based on the learnings from \S\ref{sec:content-complexity}, we use Mix Equal in letter complexity as the default setting, and pure 2-second only for efficiency concern. 
To study caption quality in isolation from visual content, we degrade captions along two controlled axes while keeping the underlying videos the same.
We define \textbf{precision} $\alpha \in [0,1]$ as the fraction of stated attributes
that are correct, and \textbf{recall} $\beta \in [0,1]$ as the fraction of true attributes that are mentioned.
When $\alpha{=}0.7$, 30\% of claims are replaced with random incorrect values, simulating hallucination.
When $\beta{=}0.3$, only 30\% of attributes appear in the caption, simulating omission.
We train separate models on a grid of 7 precision values, $\alpha \in \{1.0, 0.99, 0.95, 0.9, 0.7, 0.5, 0.3\}$, and 5 recall values, $\beta \in \{1.0, 0.9, 0.7, 0.5, 0.3\}$, for a total of 35 conditions. Training the full grid requires 35 separate models, one per condition. To keep this tractable, we train each on a 50K subset (2-second, equal-complexity mix) rather than the full 150K dataset, using a uniform 200-epoch budget set large enough for even the slowest condition to converge, so all conditions are compared at a fully-trained point. 
For each training sample, corruption is applied independently per attribute: with probability $1{-}\alpha$ a mentioned attribute is replaced with a random incorrect value, and with probability $1{-}\beta$ a ground-truth attribute is dropped from the caption.

\figref{fig:precision_recall} summarizes the result: panels~(a) and~(b) degrade one axis at a time, plotting FG~PSNR and color accuracy as precision (red) or recall (blue) is corrupted while the other axis is held uncorrupted, and panel~(c) reports the per-attribute precision advantage. All results are at CFG=1.0 on the fixed 6K held-out evaluation set described in \S\ref{sec:evaluation}. We have the following observations.

\myparagraph{Both axes degrade pixel fidelity, and the drop is a sharp cliff.}
As \figref{fig:fg_psnr} shows, FG~PSNR falls along both axes: reducing recall from $\beta{=}1.0$ to $0.3$ costs 7.6\,dB (12.3 to 4.8), and reducing precision from $\alpha{=}1.0$ to $0.3$ costs 8.1\,dB (12.3 to 4.3). Both matter because FG~PSNR captures semantic correctness alongside visual quality: wrong and missing attributes both increase pixel error against ground truth. Crucially, the precision loss is not gradual but concentrated at high $\alpha$: 5.8 of those 8.1\,dB occur between $\alpha{=}0.99$ and $\alpha{=}0.95$ (12.4 to 6.6\,dB), the steep initial drop of the red curve and a cliff from only a 4\% change in precision. A safe zone exists only at $\alpha \geq 0.99$ with $\beta{=}1.0$; the full $\alpha \times \beta$ heatmaps and per-complexity breakdowns in the appendix (\figref{fig:caption_additional}, \figref{fig:per-attr-direction}--\ref{fig:per-attr-fg-psnr}) confirm the same threshold.

\myparagraph{Precision is the primary driver of attribute controllability.}
\figref{fig:color} shows color accuracy collapsing to near chance once $\alpha \leq 0.5$: at $\beta{=}1.0$ it drops from 0.95 at $\alpha{=}1.0$ to 0.21 at $\alpha{=}0.5$ and 0.08 at $\alpha{=}0.3$, making color the most sensitive attribute to precision.
This precision dominance is not specific to color. \figref{fig:precision_advantage} reports the precision advantage, the average gain from assigning the higher quality level to precision rather than recall at a fixed corruption budget, and it is positive for all four attributes (color, direction, size, and speed). In other words, whenever caption quality is limited, spending it on correctness rather than coverage improves every attribute we measure; the per-pair detail is in Appendix~\figref{fig:swap_scatter}.
In short, \emph{mentioning} an attribute helps pixel fidelity, but stating it \emph{correctly} is what determines per-attribute controllability.

\myparagraph{\color{burntorange} Takeaways.}
Both precision ($\alpha$) and recall ($\beta$) hurt generation quality, but precision is the more damaging axis: low precision degrades pixel fidelity more sharply than low recall and, unlike recall, also collapses per-attribute accuracy. When forced to choose, \eg, under a limited budget, correctness beats completeness: for practitioners curating training data, verifying that captioned attributes are accurate should take priority over maximizing caption length or coverage.

\subsection{Caption Quality and Compute Efficiency}
\label{sec:caption-compute}

The previous section showed that caption corruption degrades final generation quality significantly.
Here we ask: Does caption corruption also affect training efficiency?
We measure the compute (in PFLOPs) needed to reach a fixed validation loss threshold, evaluated with ground-truth captions, to quantify how much harder it is for corrupted-caption models to learn the true data distribution.
We select the threshold as the lowest validation loss that all 35 conditions can achieve within the 200-epoch training budget, 
\begin{wrapfigure}{r}{0.45\textwidth}
    \centering
    \vspace{-0.12cm}
    \includegraphics[width=\linewidth]{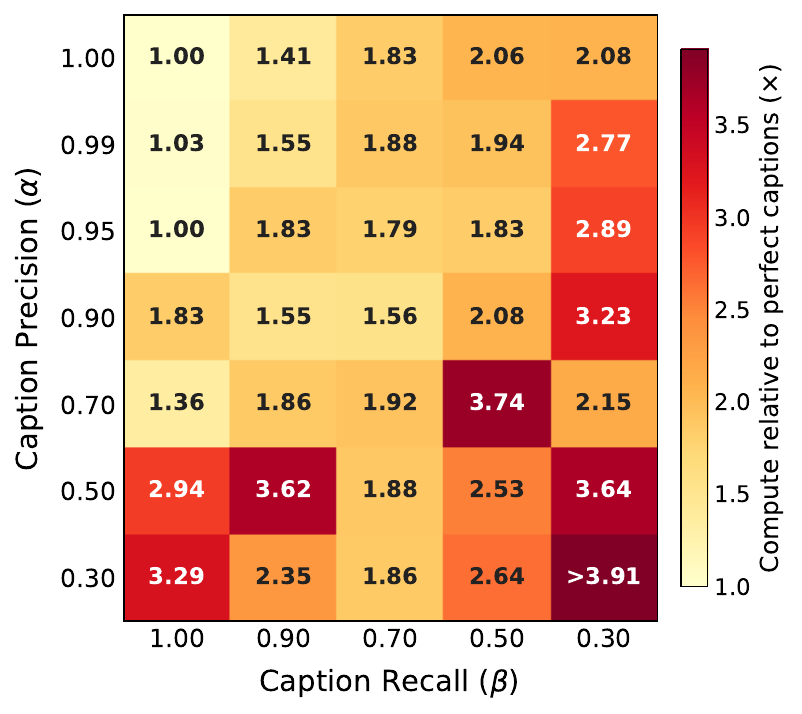}
    \vspace{-0.6cm}
    \caption{%
        Compute to reach a validation loss of 0.035 (evaluated with ground-truth captions), shown relative to the ground-truth baseline ($\alpha{=}1$, $\beta{=}1$, $66{\times}10^{3}$ PFLOPs ${=}\,1.0\times$).
        Models are trained on corrupted captions at varying precision ($\alpha$) and recall ($\beta$); higher values mean more compute to reach the same loss.
        Low recall ($\beta{=}0.3$) needs 2--4$\times$ more compute than ground-truth captions, and at $\alpha{=}0.3$, $\beta{=}0.3$ the threshold is never reached within the 200-epoch budget ($>3.9\times$).
    }
    \vspace{-1cm}
    \label{fig:caption_compute}
\end{wrapfigure}
ensuring a fair comparison across the full corruption~grid.\vspace{6pt}\\
\myparagraph{Experimental setup.}
We train the same 35 $\alpha \times \beta$ conditions from \S\ref{sec:precision-vs-recall}, but now evaluate validation loss every epoch using \emph{ground-truth} captions, measuring how well the model has learned the true data distribution regardless of its training captions.
We report the compute needed to first reach a validation loss of 0.035, normalized to the ground-truth baseline ($66{\times}10^{3}$ PFLOPs).
We select this threshold by sweeping in increments of 0.005: at 0.030 most conditions fail to converge within the training budget, making 0.035 the boundary at which the majority of settings can be compared.\vspace{6pt}\\
\myparagraph{Low recall is the most expensive axis.}
\figref{fig:caption_compute} shows a strong recall gradient: at $\beta{=}0.3$ (rightmost column), all conditions need $2.08$ to over $3.91\times$ the baseline compute, compared to $1.00$--$3.29\times$ at $\beta{=}1.0$.
At $\alpha{=}0.7$, $\beta{=}0.3$, the model reaches the threshold at $2.15\times$, slower than most $\beta{=}1.0$ conditions.
Precision also increases compute cost: at fixed $\beta{=}0.9$, reducing $\alpha$ from 1.0 to 0.3 raises cost from $1.41\times$ to $2.35\times$, with the highest cost at $\alpha{=}0.5$ ($3.62\times$).
We examine whether supervised finetuning on clean data can recover from this damage in \S\ref{sec:finetune_recovery}.
Both axes affect training efficiency, though recall shows a more consistent gradient across all precision levels. \vspace{6pt} \\
\myparagraph{\color{burntorange} Takeaways.}
Caption quality directly affects training efficiency: corrupted captions require more compute to reach a given validation loss, and in severe cases, the model never converges.
Both precision and recall increase training cost, with low recall showing a particularly consistent penalty across all precision levels.
For practitioners operating on a fixed compute budget, investing in higher-quality captions yields a larger return than training longer on noisy data. 

\begin{figure}[t]
    \centering
    \includegraphics[width=\linewidth]{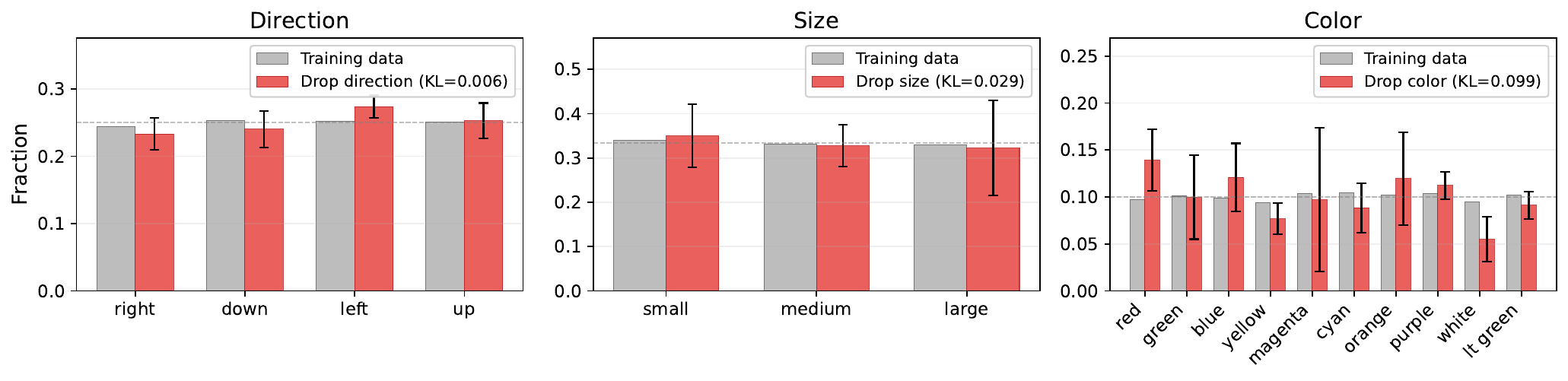}
    \vspace{-.7cm}
    \caption{%
        Drop-one marginal distributions: GT training data vs.\ generated when one attribute is dropped from otherwise-complete captions.
        Bars show mean over 4 seeds; error bars show $\pm$1 std.
        Direction stays near-uniform; color and size develop seed-dependent biases.
    }
    \label{fig:drop_one_dist}
\end{figure}
\FloatBarrier
\subsection{What Happens When Attributes Are Never Mentioned?}
\label{sec:marginal-bias}

Real-world captions rarely describe every visual attribute.
Even detailed captions cannot cover all aspects of a video: a caption might describe the main subjects and actions, but omit details like the exact colors of background objects, the precise speed of each motion, or the size relationships between elements.
Since achieving perfect recall is practically impossible, attributes are systematically omitted from training captions.
When an attribute is systematically absent from all training captions, the model must decide what to generate.
We study whether models learn to produce diverse values or collapse to a default.

\myparagraph{Experimental setup.}
We drop one attribute from otherwise-complete captions: for each target attribute (size, direction, or color), we train a model on captions that are identical to the ground-truth captions except that every mention of the target attribute is removed, while all other attributes remain fully specified and vary across videos.
We repeat each condition across 4 random seeds and evaluate at CFG\,=\,1.0.

\myparagraph{Size and color develop biases; direction is resilient.}
When one attribute is dropped from otherwise-complete captions (\figref{fig:drop_one_dist}), size develops the largest bias (mean KL\,=\,0.029), though with high seed-dependent variance (0.003--0.056).
Color also develops biases, with each seed favoring different colors (mean KL\,=\,0.099).
Direction, by contrast, stays near-uniform across all seeds (mean KL\,=\,0.006). The reason is that direction is redundantly specified by other fields: when the direction phrase is dropped, the start and end positions (and any bounce events) remain in the caption, and together they largely determine the net direction even without stating it explicitly.

\myparagraph{\color{burntorange} Takeaways.} 
Whether a missing attribute is still generated correctly depends on whether it remains recoverable from the rest of the caption. Direction is resilient because it stays pinned down by the start and end positions left in the caption when the direction phrase is dropped; size and color have no such redundant cue and instead develop biases, with each model drifting toward a default value whose choice varies unpredictably across seeds.
This matters for real data. Web captions rarely mention attributes like object size (``a cat on a table'' does not say whether the cat is large or small). Omitting such an attribute does not preserve its natural distribution; instead, each model develops its own run-dependent bias, making the attribute uncontrollable rather than reliably generated.
Similarly, captions for dance or sports videos describe the activity (``people dancing'') without specifying speed, pose, or spatial arrangement, so these attributes converge to per-model defaults.
The practical fix is to explicitly caption the attributes that matter for generation, unless they are already implied by other fields that remain in the caption.

\FloatBarrier
\section{Investigation on the Recovery from Poor Pre-training}
\label{sec:training-recipe}

The previous sections showed that caption corruption during pretraining causes lasting damage to generation quality.
We now ask whether that damage can be undone after the fact, through two mitigation strategies: classifier-free guidance (CFG) at inference time (\S\ref{sec:cfg_compensation}) and finetuning on a small set of clean captions (\S\ref{sec:finetune_recovery}).
For each, we measure how much quality is recovered and where it falls short.

\begin{figure*}[t]
    \centering
    \includegraphics[width=0.65\linewidth]{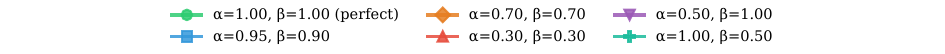}\\[2pt]
    \begin{subfigure}[t]{0.24\textwidth}
        \includegraphics[width=\linewidth]{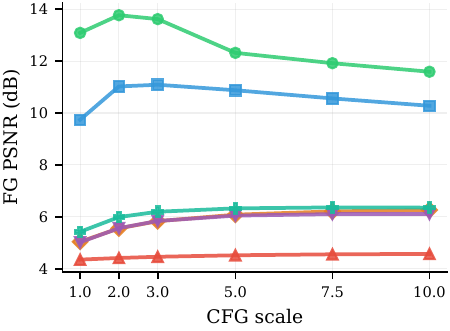}
        \caption{FG~PSNR}
        \label{fig:cfg_psnr}
    \end{subfigure}\hfill
    \begin{subfigure}[t]{0.24\textwidth}
        \includegraphics[width=\linewidth]{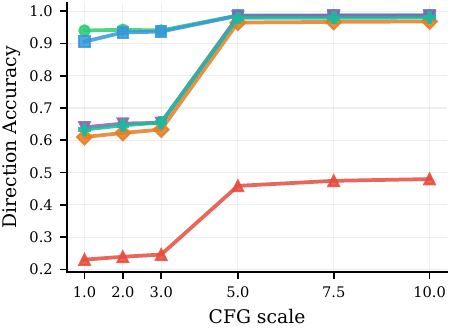}
        \caption{Direction}
        \label{fig:cfg_dir}
    \end{subfigure}\hfill
    \begin{subfigure}[t]{0.24\textwidth}
        \includegraphics[width=\linewidth]{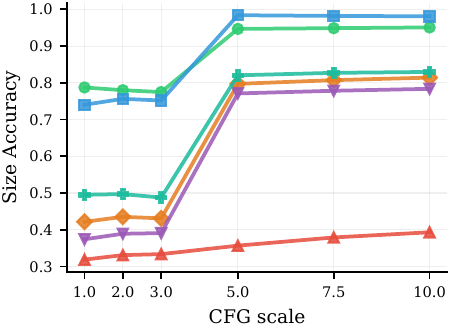}
        \caption{Size}
        \label{fig:cfg_size}
    \end{subfigure}\hfill
    \begin{subfigure}[t]{0.24\textwidth}
        \includegraphics[width=\linewidth]{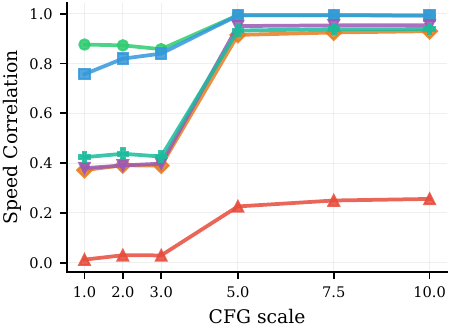}
        \caption{Speed}
        \label{fig:cfg_speed}
    \end{subfigure}
    \caption{%
        Effect of CFG scale on generation quality for models trained with varying caption corruption.
        Each line is one corruption condition averaged across 1L/2L/3L evaluation.
        Higher CFG improves attribute faithfulness but degrades pixel quality, revealing a quality--faithfulness trade-off. 
    }
    \label{fig:cfg_sweep}
\end{figure*}

\FloatBarrier
\subsection{How Does CFG Interact with Caption Quality?}
\label{sec:cfg_compensation}

\myparagraph{Experimental setup.}
Classifier-free guidance (CFG)~\cite{ho2022classifier} amplifies the conditional signal at inference time and is widely used in practice.
We study how CFG interacts with training data quality by evaluating every model from the caption corruption grid in \S\ref{sec:precision-vs-recall} at six CFG scales from 1.0 to 10.0, keeping all other inference settings fixed at 50 steps and 2000 samples.
\figref{fig:cfg_sweep} shows four metrics, FG~PSNR, direction, size, and speed, averaged across complexity levels for six representative conditions spanning ground-truth to severe corruption. We omit color from the figure to avoid clutter; as the most precision-sensitive attribute (\S\ref{sec:precision-vs-recall}), its relevant values under CFG are reported inline below.

\myparagraph{With good captions, CFG has a sweet spot.}
For the model trained on ground-truth captions, FG~PSNR peaks at CFG\,=\,2 (12.6\,dB) then declines to 11.5\,dB at CFG\,=\,10 (\figref{fig:cfg_psnr}).
The mild corruption condition ($\alpha{=}0.95$, $\beta{=}0.9$) shows a similar peak at CFG\,=\,2--3.
For the moderate corruption model ($\alpha{=}0.7$, $\beta{=}0.7$), direction accuracy rises from 0.94 to 0.97 at CFG\,=\,5 (\figref{fig:cfg_dir}).
This reveals a sweet spot around CFG\,=\,2--3 that balances pixel fidelity and attribute faithfulness when captions are good.

\myparagraph{CFG helps accuracy, but cannot recover pixel quality.}
For moderately corrupted models ($\alpha{=}0.7$, $\beta{=}0.7$), high CFG dramatically improves per-attribute accuracy: direction rises from 0.94 to 0.97, and size from 0.66 to 0.80 at CFG\,=\,5+ (\figref{fig:cfg_dir}, \figref{fig:cfg_size}).
However, FG~PSNR barely improves from 5.0 to 6.3\,dB, against a gap of 7.3\,dB to the model trained on ground-truth captions (\figref{fig:cfg_psnr}).
The pixel-quality damage from bad captions is irrecoverable through CFG, even as per-attribute accuracy converges toward clean-model levels.
This split occurs because CFG can steer the model toward correct categorical decisions (direction, size), but it cannot undo the corrupted internal representations that degrade rendering quality. 

\myparagraph{Severe corruption is beyond CFG's reach.}
At $\alpha{=}0.3$, $\beta{=}0.3$, neither pixel quality nor accuracy improves meaningfully with CFG.
FG~PSNR varies by less than 0.3\,dB across the full range (\figref{fig:cfg_psnr}), and direction accuracy reaches only 0.49 at CFG\,=\,10 (\figref{fig:cfg_dir}).
Color accuracy is already near chance (0.07 at CFG\,=\,1) and shows no recovery, since 70\% of color claims are wrong and stronger guidance pushes harder toward incorrect values.
When the majority of caption information is incorrect, CFG amplifies errors as readily as the correct signal.

\begin{figure}[t]
    \centering
    \includegraphics[width=0.95\linewidth]{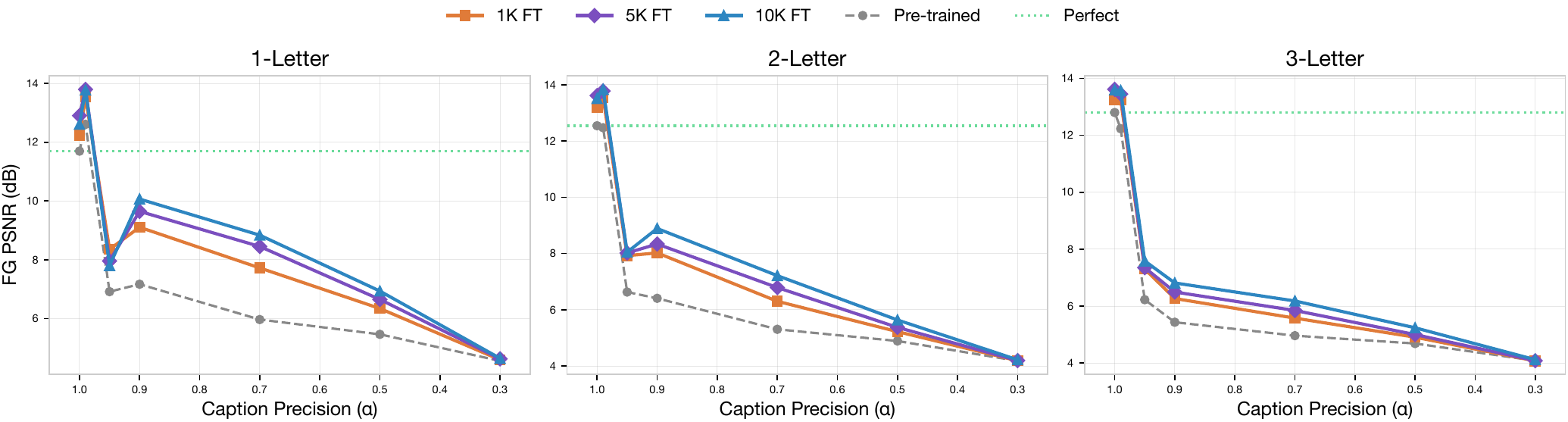}
    \caption{%
        Finetuning recovery for FG~PSNR at $\beta{=}1.0$, CFG\,=\,1.0.
        Raw FG~PSNR vs.\ caption precision $\alpha$.
        Finetuned models partially recover for $\alpha \geq 0.7$ but show no improvement for $\alpha \leq 0.5$.
        10K outperforms 5K, which outperforms 1K, with diminishing returns. 
    }
    \label{fig:finetune_recovery}
\end{figure}

\myparagraph{Precision and recall interact differently with CFG.}
Comparing the $\alpha$-only ($\alpha{=}0.5$, $\beta{=}1.0$) and $\beta$-only ($\alpha{=}1.0$, $\beta{=}0.5$) conditions isolates the two caption failure modes.
Both have similar FG~PSNR (5.0 vs.\ 5.4\,dB at CFG\,=\,1), but their accuracy profiles diverge under CFG.
The $\beta$-only model retains higher color accuracy (0.39 at CFG\,=\,1) because all stated attributes are correct, \ie, they are merely incomplete.
The $\alpha$-only model starts lower (0.21) and shows less recovery, because half of its color claims are wrong and CFG amplifies those errors.
This confirms that omission ($\beta$) is a more benign failure mode than hallucination ($\alpha$). Models trained on incomplete but accurate captions are better on CFG than models trained on complete but inaccurate ones.

\myparagraph{\color{burntorange} Takeaways.}
CFG reveals a clear asymmetry that it can recover per-attribute accuracy for moderate corruption but cannot restore pixel-level fidelity at any corruption level.
With good captions, a sweet spot around CFG\,=\,2--3 optimizes the trade-off between quality and faithfulness.
With bad captions, no CFG setting substitutes for clean training data.
Between the two failure modes, omission (low $\beta$) is more amenable to CFG correction than hallucination (low $\alpha$). 

\FloatBarrier
\subsection{Can Finetuning Recover from Caption Corruption?}
\label{sec:finetune_recovery}

\myparagraph{Experimental setup.}
Since neither CFG nor the model's own robustness can compensate for poor training captions, we ask whether finetuning with correct data can undo the damage.
We take each caption-corrupted model from \S\ref{sec:precision-vs-recall} and finetune it for 200 epochs with ground-truth captions at a reduced learning rate of $10^{-4}$, down from $5\times10^{-4}$, using 1K, 5K, or 10K stratified samples balanced across complexity levels.
We evaluate at finetuning epochs 100 and 200.
\figref{fig:finetune_recovery} shows FG~PSNR at $\beta{=}1.0$ across $\alpha$ values.

\myparagraph{Mild corruption is partially recoverable.}
Models trained at $\alpha \geq 0.7$ recover a meaningful fraction of lost FG~PSNR, but not all.
At $\alpha{=}0.9$, finetuning with 5K samples raises 1L FG~PSNR from 7.2 to 9.6\,dB, a 55\% recovery toward the 11.7\,dB baseline.
At $\alpha{=}0.7$, 1L recovers 43\%, 2L recovers 21\%, and 3L only 11\%.

\myparagraph{Severe corruption leaves permanent damage.}
At $\alpha \leq 0.5$, finetuning yields negligible improvement.
At $\alpha{=}0.3$, FG~PSNR barely improves: 1L stays at 4.6\,dB versus 4.6\,dB pretrained, essentially unchanged; see \figref{fig:finetune_recovery}.
Severely corrupted pretraining learns incorrect associations that even 10K correct examples cannot overwrite.

\begin{figure}[t]
    \centering
    \begin{subfigure}[t]{0.77\linewidth}
        \includegraphics[width=\linewidth]{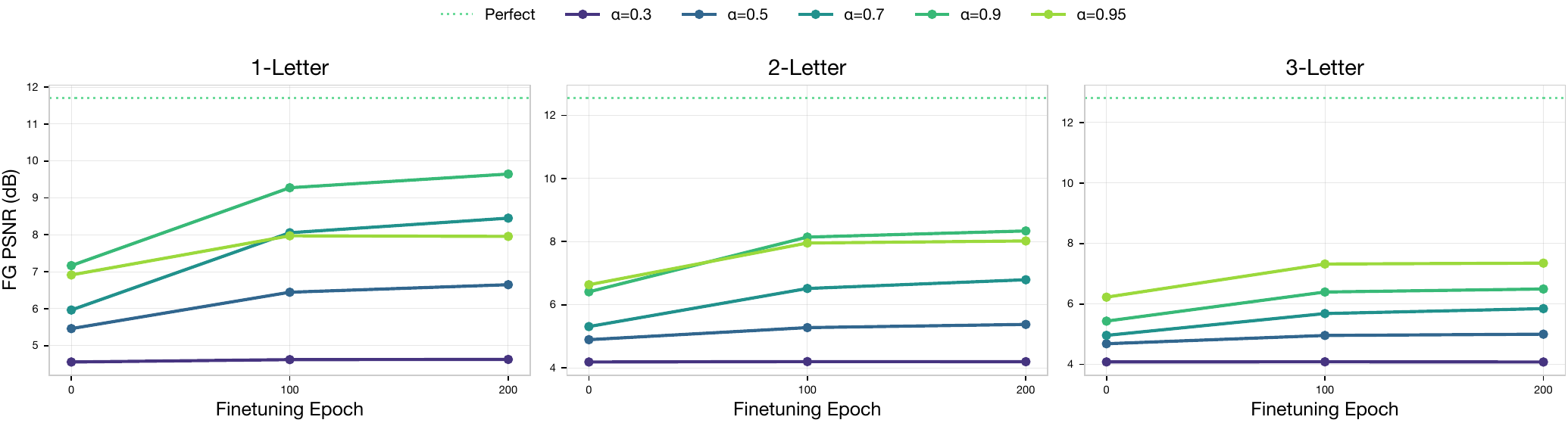}
        \caption{Recovery over finetuning epochs}
        \label{fig:ft_recovery_curves}
    \end{subfigure}\hfill
    \begin{subfigure}[t]{0.2\linewidth}
        \includegraphics[width=\linewidth]{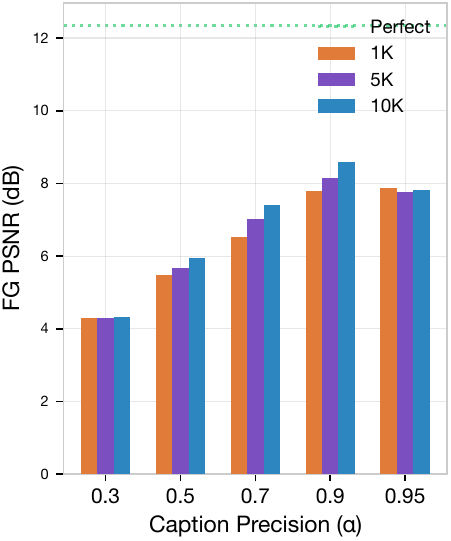}
        \caption{1K vs.\ 5K finetuning}
        \label{fig:ft_1k_vs_5k}
    \end{subfigure}
    \caption{%
        (a) FG~PSNR recovery over finetuning epochs. Most gains occur within the first 100 epochs; severe corruption ($\alpha{=}0.3$) shows no recovery.
        (b) 10K finetuning outperforms 5K and 1K for FG~PSNR, with the largest margin at moderate corruption.
    }
    \label{fig:ft_recovery_vs_alpha}
\end{figure}

\myparagraph{Recovery is harder for compositional scenes.}
A clear complexity gradient emerges: 1L scenes consistently achieve the highest recovery, followed by 2L, then 3L; see \figref{fig:ft_recovery_vs_alpha}.
At $\alpha{=}0.7$, 1L reaches 43\% recovery while 3L reaches only 11\% with 5K samples.
Multi-object scenes require learning inter-object relationships like relative positions and independent motions, which are harder to reacquire from limited data.

\begin{figure}[t]
    \centering
    \includegraphics[width=0.95\linewidth]{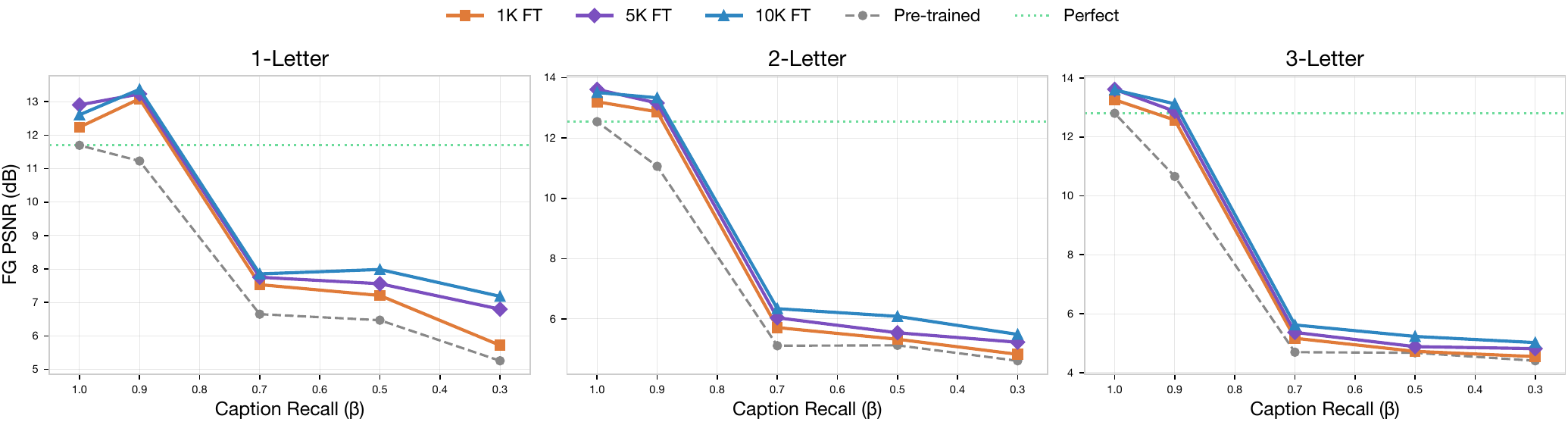}
    \caption{%
        Finetuning recovery for FG~PSNR vs.\ caption recall $\beta$ at fixed $\alpha{=}1.0$, CFG\,=\,1.0.
        Mild recall corruption ($\beta{=}0.9$) is fully recoverable, with finetuned FG~PSNR exceeding the ground-truth baseline.
        At lower recall ($\beta \leq 0.7$), finetuning recovers only 1--2\,dB regardless of data size, leaving a large gap to the baseline.
    }
    \label{fig:ft_recovery_beta}
\end{figure}

\myparagraph{More data helps, but with diminishing returns.}
Finetuning with more data consistently helps, as shown in \figref{fig:ft_1k_vs_5k}: 10K outperforms 5K, which outperforms 1K, with the largest gains in the moderate corruption regime at $\alpha \in [0.7, 0.95]$.
For severe corruption at $\alpha \leq 0.5$, even 10K samples cannot help. The damage is structural, not data-limited.

\myparagraph{Recall corruption is far more recoverable than precision.}
Recall corruption behaves very differently from precision (\figref{fig:ft_recovery_beta}, alongside the precision results in \figref{fig:finetune_recovery}).
At $\beta{=}0.9$, finetuning fully recovers FG~PSNR, matching or even exceeding the ground-truth baseline.
Recovery still falls off at lower recall (under 20\% for $\beta \leq 0.7$), but mild omission is essentially free to repair.
This mirrors the pretraining finding (\S\ref{sec:precision-vs-recall}): a model trained on incomplete but accurate captions has learned nothing wrong, so clean data simply fills in the gaps, whereas hallucinated associations must first be unlearned.

\myparagraph{\color{burntorange} Takeaways.}
Finetuning on correct data provides modest FG~PSNR recovery for mild corruption ($\alpha \geq 0.7$) but cannot rescue severe damage ($\alpha \leq 0.5$), and most of the recovery happens within the first 100 epochs (\figref{fig:ft_recovery_curves}).
The pretraining asymmetry persists: omission (low $\beta$) is much easier to repair than hallucination (low $\alpha$), since the model has not learned wrong associations.

\section{Conclusion}
\label{sec:conclusion}

We introduce \textbf{Moving Alphabet}, a controlled testbed for studying how training data affects text-to-video generation. Our experiments yield three findings: balanced and diverse content complexity and clip duration are important for generalization; caption quality is crucial for both model performance and training efficiency; and post-training remedies such as classifier-free guidance and fine-tuning provide only partial recovery from poor pre-training data. These results collectively recommend that practitioners prioritize balanced data curation and high-quality captions at the pre-training stage. This work also opens up a few interesting future directions, including how to measure diversity in real-world scenarios, and conditioning on non-textual modalities such as spatial layouts~\cite{reve2026layoutbet}. In summary, we hope this work provides a clear demonstration of the importance of training data in text-to-video generation and encourages efforts on data curation and quality improvement in the community.

\clearpage
\newpage
\bibliographystyle{assets/plainnat}
\bibliography{paper}

\begin{thebibliography}{45}
\providecommand{\natexlab}[1]{#1}
\providecommand{\url}[1]{\texttt{#1}}
\expandafter\ifx\csname urlstyle\endcsname\relax
  \providecommand{\doi}[1]{doi: #1}\else
  \providecommand{\doi}{doi: \begingroup \urlstyle{rm}\Url}\fi

\bibitem[Allen-Zhu and Li(2024)]{allenzhu2024physics}
Zeyuan Allen-Zhu and Yuanzhi Li.
\newblock Physics of language models: Part 3.1, knowledge storage and extraction.
\newblock In \emph{Proc. ICML}, 2024.

\bibitem[Bar-Tal et~al.(2024)Bar-Tal, Chefer, Tov, Herrmann, Paiss, Zada, Ephrat, Hur, Li, Michaeli, Wang, Sun, Dekel, and Mosseri]{bartal2024lumiere}
Omer Bar-Tal, Hila Chefer, Omer Tov, Charles Herrmann, Roni Paiss, Shiran Zada, Ariel Ephrat, Junhwa Hur, Yuanzhen Li, Tomer Michaeli, Oliver Wang, Deqing Sun, Tali Dekel, and Inbar Mosseri.
\newblock Lumiere: A space-time diffusion model for video generation.
\newblock \emph{arXiv preprint arXiv:2401.12945}, 2024.

\bibitem[Bengio et~al.(2009)Bengio, Louradour, Collobert, and Weston]{bengio2009curriculum}
Yoshua Bengio, J{\'e}r{\^o}me Louradour, Ronan Collobert, and Jason Weston.
\newblock Curriculum learning.
\newblock In \emph{Proc. ICML}, 2009.

\bibitem[Betker et~al.(2023)Betker, Goh, Jing, Brooks, Wang, Li, Ouyang, Zhuang, Lee, Guo, et~al.]{betker2023dalle3}
James Betker, Gabriel Goh, Li~Jing, Tim Brooks, Jianfeng Wang, Linjie Li, Long Ouyang, Juntang Zhuang, Joyce Lee, Yufei Guo, et~al.
\newblock Improving image generation with better captions.
\newblock 2023.
\newblock OpenAI technical report.

\bibitem[Blattmann et~al.(2023)Blattmann, Dockhorn, Kulal, Mendelevitch, Kilian, Lorber, Levi, English, Voleti, Letts, et~al.]{blattmann2023stablevideo}
Andreas Blattmann, Tim Dockhorn, Sumith Kulal, Daniel Mendelevitch, Maciej Kilian, Dominik Lorber, Yam Levi, Zion English, Vikram Voleti, Adam Letts, et~al.
\newblock Stable video diffusion: Scaling latent video diffusion models to large datasets.
\newblock \emph{arXiv preprint arXiv:2311.15127}, 2023.

\bibitem[Chen et~al.(2024{\natexlab{a}})Chen, Zhang, Cun, Xia, Wang, Weng, and Shan]{chen2024videocrafter2}
Haoxin Chen, Yong Zhang, Xiaodong Cun, Menghan Xia, Xintao Wang, Chao Weng, and Ying Shan.
\newblock {VideoCrafter2}: Overcoming data limitations for high-quality video diffusion models.
\newblock \emph{arXiv preprint arXiv:2401.09047}, 2024{\natexlab{a}}.

\bibitem[Chen et~al.(2024{\natexlab{b}})Chen, Wei, Li, Dong, Zhang, Zang, Chen, Duan, Lin, Tang, Yuan, Qiao, Lin, Zhao, and Wang]{chen2025sharegpt4video}
Lin Chen, Xilin Wei, Jinsong Li, Xiaoyi Dong, Pan Zhang, Yuhang Zang, Zehui Chen, Haodong Duan, Bin Lin, Zhenyu Tang, Li~Yuan, Yu~Qiao, Dahua Lin, Feng Zhao, and Jiaqi Wang.
\newblock {ShareGPT4Video}: Improving video understanding and generation with better captions.
\newblock \emph{arXiv preprint arXiv:2406.04325}, 2024{\natexlab{b}}.

\bibitem[Chen et~al.(2024{\natexlab{c}})Chen, Siarohin, Menapace, Deyneka, Chao, Jeon, Fang, Lee, Ren, Yang, and Tulyakov]{chen2024panda70m}
Tsai-Shien Chen, Aliaksandr Siarohin, Willi Menapace, Ekaterina Deyneka, Hsiang-wei Chao, Byung~Eun Jeon, Yuwei Fang, Hsin-Ying Lee, Jian Ren, Ming-Hsuan Yang, and Sergey Tulyakov.
\newblock {Panda-70M}: Captioning 70{M} videos with multiple cross-modality teachers.
\newblock \emph{arXiv preprint arXiv:2402.19479}, 2024{\natexlab{c}}.

\bibitem[Comanici et~al.(2025)Comanici, Bieber, Schaekermann, Pasupat, Sachdeva, Dhillon, Blistein, Ram, Zhang, Rosen, et~al.]{comanici2025gemini}
Gheorghe Comanici, Eric Bieber, Mike Schaekermann, Ice Pasupat, Noveen Sachdeva, Inderjit Dhillon, Marcel Blistein, Ori Ram, Dan Zhang, Evan Rosen, et~al.
\newblock Gemini 2.5: Pushing the frontier with advanced reasoning, multimodality, long context, and next generation agentic capabilities.
\newblock \emph{arXiv preprint arXiv:2507.06261}, 2025.

\bibitem[Eldan and Li(2023)]{eldan2023tinystories}
Ronen Eldan and Yuanzhi Li.
\newblock {TinyStories}: How small can language models be and still speak coherent {English}?
\newblock \emph{arXiv preprint arXiv:2305.07759}, 2023.

\bibitem[Esser et~al.(2024)Esser, Kulal, Blattmann, Entezari, M\"{u}ller, Saini, Levi, Lorenz, Sauer, Boesel, Podell, Dockhorn, English, and Rombach]{mmdit}
Patrick Esser, Sumith Kulal, Andreas Blattmann, Rahim Entezari, Jonas M\"{u}ller, Harry Saini, Yam Levi, Dominik Lorenz, Axel Sauer, Frederic Boesel, Dustin Podell, Tim Dockhorn, Zion English, and Robin Rombach.
\newblock Scaling rectified flow transformers for high-resolution image synthesis.
\newblock In \emph{Proc. ICML}, 2024.

\bibitem[Gadre et~al.(2023)Gadre, Ilharco, Fang, Hayase, Smyrnis, Nguyen, Marten, Wortsman, Ghosh, Zhang, et~al.]{datacomp}
Samir~Yitzhak Gadre, Gabriel Ilharco, Alex Fang, Jonathan Hayase, Georgios Smyrnis, Thao Nguyen, Ryan Marten, Mitchell Wortsman, Dhruba Ghosh, Jieyu Zhang, et~al.
\newblock {DataComp}: In search of the next generation of multimodal datasets.
\newblock In \emph{Proc. NeurIPS}, 2023.

\bibitem[Gao et~al.(2020)Gao, Biderman, Black, Golding, Hoppe, Foster, Phang, He, Thite, Nabeshima, Presser, and Leahy]{gao2020pile}
Leo Gao, Stella Biderman, Sid Black, Laurence Golding, Travis Hoppe, Charles Foster, Jason Phang, Horace He, Anish Thite, Noa Nabeshima, Shawn Presser, and Connor Leahy.
\newblock The {Pile}: An 800{GB} dataset of diverse text for language modeling.
\newblock \emph{arXiv preprint arXiv:2101.00027}, 2020.

\bibitem[Ho and Salimans(2022)]{ho2022classifier}
Jonathan Ho and Tim Salimans.
\newblock Classifier-free diffusion guidance.
\newblock \emph{arXiv preprint arXiv:2207.12598}, 2022.

\bibitem[Ho et~al.(2020)Ho, Jain, and Abbeel]{ho2020denoising}
Jonathan Ho, Ajay Jain, and Pieter Abbeel.
\newblock Denoising diffusion probabilistic models.
\newblock In \emph{Proc. NeurIPS}, 2020.

\bibitem[Ho et~al.(2022)Ho, Chan, Saharia, Whang, Gao, Gritsenko, Kingma, Poole, Norouzi, Fleet, and Salimans]{ho2022imagenvideo}
Jonathan Ho, William Chan, Chitwan Saharia, Jay Whang, Ruiqi Gao, Alexey Gritsenko, Diederik~P Kingma, Ben Poole, Mohammad Norouzi, David~J Fleet, and Tim Salimans.
\newblock Imagen video: High definition video generation with diffusion models.
\newblock \emph{arXiv preprint arXiv:2210.02303}, 2022.

\bibitem[Huang et~al.(2023)Huang, Sun, Xie, Li, and Liu]{huang2023t2icompbench}
Kaiyi Huang, Kaiyue Sun, Enze Xie, Zhenguo Li, and Xihui Liu.
\newblock {T2I-CompBench}: A comprehensive benchmark for open-world compositional text-to-image generation.
\newblock \emph{arXiv preprint arXiv:2307.06350}, 2023.

\bibitem[Johnson et~al.(2017)Johnson, Hariharan, van~der Maaten, Fei-Fei, Zitnick, and Girshick]{johnson2017clevr}
Justin Johnson, Bharath Hariharan, Laurens van~der Maaten, Li~Fei-Fei, C~Lawrence Zitnick, and Ross Girshick.
\newblock {CLEVR}: A diagnostic dataset for compositional language and elementary visual reasoning.
\newblock In \emph{Proc. CVPR}, 2017.

\bibitem[Kaplan et~al.(2020)Kaplan, McCandlish, Henighan, Brown, Chess, Child, Gray, Radford, Wu, and Amodei]{kaplan2020scaling}
Jared Kaplan, Sam McCandlish, Tom Henighan, Tom~B Brown, Benjamin Chess, Rewon Child, Scott Gray, Alec Radford, Jeffrey Wu, and Dario Amodei.
\newblock Scaling laws for neural language models.
\newblock \emph{arXiv preprint arXiv:2001.08361}, 2020.

\bibitem[Kim and Mnih(2018)]{kim2018shapes3d}
Hyunjik Kim and Andriy Mnih.
\newblock Disentangling by factorising.
\newblock In \emph{Proc. ICML}, 2018.

\bibitem[Kong et~al.(2024)Kong, Tian, Zhang, Min, Dai, Zhou, Xu, Pang, Liu, et~al.]{kong2024hunyuanvideo}
Weijie Kong, Qi~Tian, Zijian Zhang, Rox Min, Zuozhuo Dai, Jin Zhou, Jiangfeng Xu, Bohao Pang, Hao Liu, et~al.
\newblock {HunyuanVideo}: A systematic framework for large video generative models.
\newblock \emph{arXiv preprint arXiv:2412.03603}, 2024.

\bibitem[Lipman et~al.(2023)Lipman, Chen, Ben-Hamu, Nickel, and Le]{lipman2023flow}
Yaron Lipman, Ricky T.~Q. Chen, Heli Ben-Hamu, Maximilian Nickel, and Matthew Le.
\newblock Flow matching for generative modeling.
\newblock In \emph{Proc. ICLR}, 2023.

\bibitem[Liu et~al.(2024)Liu, Li, Li, Li, Zhang, Shen, and Lee]{liu2024llavanext}
Haotian Liu, Chunyuan Li, Yuheng Li, Bo~Li, Yuanhan Zhang, Sheng Shen, and Yong~Jae Lee.
\newblock {LLaVA-NeXT}: Improved reasoning, {OCR}, and world knowledge.
\newblock 2024.
\newblock Blog post.

\bibitem[Ma et~al.(2024)Ma, Wang, Jia, Chen, Liu, Li, Chen, and Qiao]{ma2024latte}
Xin Ma, Yaohui Wang, Gengyun Jia, Xinyuan Chen, Ziwei Liu, Yuan-Fang Li, Cunjian Chen, and Yu~Qiao.
\newblock Latte: Latent diffusion transformer for video generation.
\newblock \emph{arXiv preprint arXiv:2401.03048}, 2024.

\bibitem[Matthey et~al.(2017)Matthey, Higgins, Hassabis, and Lerchner]{matthey2017dsprites}
Loic Matthey, Irina Higgins, Demis Hassabis, and Alexander Lerchner.
\newblock d{S}prites: Disentanglement testing sprites dataset.
\newblock 2017.
\newblock https://github.com/deepmind/dsprites-dataset/.

\bibitem[Muennighoff et~al.(2023)Muennighoff, Rush, Barak, Le~Scao, Piktus, Tazi, Pyysalo, Wolf, and Raffel]{muennighoff2024scaling}
Niklas Muennighoff, Alexander~M Rush, Boaz Barak, Teven Le~Scao, Aleksandra Piktus, Nouamane Tazi, Sampo Pyysalo, Thomas Wolf, and Colin Raffel.
\newblock Scaling data-constrained language models.
\newblock \emph{arXiv preprint arXiv:2305.16264}, 2023.

\bibitem[OpenAI(2024)]{sora}
OpenAI.
\newblock Sora: Creating video from text.
\newblock 2024.
\newblock Technical report.

\bibitem[Peebles and Xie(2023)]{peebles2023scalable}
William Peebles and Saining Xie.
\newblock Scalable diffusion models with transformers.
\newblock In \emph{Proc. ICCV}, 2023.

\bibitem[Polyak et~al.(2024)Polyak, Zohar, Brown, Tjandra, Sinha, Lee, Vyas, Shi, Ma, Chuang, et~al.]{polyak2024moviegen}
Adam Polyak, Amit Zohar, Andrew Brown, Andros Tjandra, Animesh Sinha, Ann Lee, Apoorv Vyas, Bowen Shi, Chih-Yao Ma, Ching-Yao Chuang, et~al.
\newblock Movie {Gen}: A cast of media foundation models.
\newblock \emph{arXiv preprint arXiv:2410.13720}, 2024.

\bibitem[Porter and Duff(1984)]{porter1984compositing}
Thomas Porter and Tom Duff.
\newblock Compositing digital images.
\newblock In \emph{Proc. SIGGRAPH}, 1984.

\bibitem[Radford et~al.(2021)Radford, Kim, Hallacy, Ramesh, Goh, Agarwal, Sastry, Askell, Mishkin, Clark, Krueger, and Sutskever]{radford2021clip}
Alec Radford, Jong~Wook Kim, Chris Hallacy, Aditya Ramesh, Gabriel Goh, Sandhini Agarwal, Girish Sastry, Amanda Askell, Pamela Mishkin, Jack Clark, Gretchen Krueger, and Ilya Sutskever.
\newblock Learning transferable visual models from natural language supervision.
\newblock In \emph{Proc. ICML}, 2021.

\bibitem[Raffel et~al.(2020)Raffel, Shazeer, Roberts, Lee, Narang, Matena, Zhou, Li, and Liu]{raffel2020t5}
Colin Raffel, Noam Shazeer, Adam Roberts, Katherine Lee, Sharan Narang, Michael Matena, Yanqi Zhou, Wei Li, and Peter~J. Liu.
\newblock Exploring the limits of transfer learning with a unified text-to-text transformer.
\newblock \emph{JMLR}, 2020.

\bibitem[{Reve Team}(2026)]{reve2026layoutbet}
{Reve Team}.
\newblock The layout bet, June 2026.
\newblock \url{https://reve.com}.
\newblock Accessed: 2026-06-03.

\bibitem[Rombach et~al.(2022)Rombach, Blattmann, Lorenz, Esser, and Ommer]{rombach2022high}
Robin Rombach, Andreas Blattmann, Dominik Lorenz, Patrick Esser, and Bj{\"o}rn Ommer.
\newblock High-resolution image synthesis with latent diffusion models.
\newblock In \emph{Proc. CVPR}, 2022.

\bibitem[Schuhmann et~al.(2022)Schuhmann, Beaumont, Vencu, Gordon, Wightman, Cherti, Coombes, Katta, Mullis, Wortsman, et~al.]{schuhmann2022laion5b}
Christoph Schuhmann, Romain Beaumont, Richard Vencu, Cade Gordon, Ross Wightman, Mehdi Cherti, Theo Coombes, Aarush Katta, Clayton Mullis, Mitchell Wortsman, et~al.
\newblock {LAION}-5{B}: An open large-scale dataset for training next generation image-text models.
\newblock In \emph{Proc. NeurIPS}, 2022.

\bibitem[Singer et~al.(2022)Singer, Polyak, Hayes, Yin, An, Zhang, Hu, Yang, Ashual, Gafni, Parikh, Gupta, and Taigman]{singer2022makeavideo}
Uriel Singer, Adam Polyak, Thomas Hayes, Xi~Yin, Jie An, Songyang Zhang, Qiyuan Hu, Harry Yang, Oron Ashual, Oran Gafni, Devi Parikh, Sonal Gupta, and Yaniv Taigman.
\newblock {Make-A-Video}: Text-to-video generation without text-video data.
\newblock \emph{arXiv preprint arXiv:2209.14792}, 2022.

\bibitem[Team et~al.(2025)Team, Du, Yin, Xing, Qu, Wang, Chen, Zhang, Du, Wei, et~al.]{team2025kimi}
Kimi Team, Angang Du, Bohong Yin, Bowei Xing, Bowen Qu, Bowen Wang, Cheng Chen, Chenlin Zhang, Chenzhuang Du, Chu Wei, et~al.
\newblock {Kimi-VL} technical report.
\newblock \emph{arXiv preprint arXiv:2504.07491}, 2025.

\bibitem[Unterthiner et~al.(2019)Unterthiner, van Steenkiste, Kurach, Marinier, Michalski, and Gelly]{unterthiner2019fvd}
Thomas Unterthiner, Sjoerd van Steenkiste, Karol Kurach, Rapha{\"e}l Marinier, Marcin Michalski, and Sylvain Gelly.
\newblock {FVD}: A new metric for video generation.
\newblock 2019.

\bibitem[Vaswani et~al.(2017)Vaswani, Shazeer, Parmar, Uszkoreit, Jones, Gomez, Kaiser, and Polosukhin]{vaswani2017attention}
Ashish Vaswani, Noam Shazeer, Niki Parmar, Jakob Uszkoreit, Llion Jones, Aidan~N. Gomez, {\L}ukasz Kaiser, and Illia Polosukhin.
\newblock Attention is all you need.
\newblock In \emph{Proc. NeurIPS}, 2017.

\bibitem[Wang et~al.(2023)Wang, He, Li, Li, Yu, Ma, Li, Chen, Chen, Wang, et~al.]{wang2024internvid}
Yi~Wang, Yinan He, Yizhuo Li, Kunchang Li, Jiashuo Yu, Xin Ma, Xinhao Li, Guo Chen, Xinyuan Chen, Yaohui Wang, et~al.
\newblock {InternVid}: A large-scale video-text dataset for multimodal understanding and generation.
\newblock \emph{arXiv preprint arXiv:2307.06942}, 2023.

\bibitem[Xie et~al.(2023)Xie, Santurkar, Ma, and Liang]{xie2023doremi}
Sang~Michael Xie, Shibani Santurkar, Tengyu Ma, and Percy Liang.
\newblock {DoReMi}: Optimizing data mixtures speeds up language model pretraining.
\newblock In \emph{Proc. NeurIPS}, 2023.

\bibitem[Yang et~al.(2025)Yang, Li, Yang, Zhang, Hui, Zheng, Yu, Gao, Huang, Lv, et~al.]{yang2025qwen3}
An~Yang, Anfeng Li, Baosong Yang, Beichen Zhang, Binyuan Hui, Bo~Zheng, Bowen Yu, Chang Gao, Chengen Huang, Chenxu Lv, et~al.
\newblock Qwen3 technical report.
\newblock \emph{arXiv preprint arXiv:2505.09388}, 2025.

\bibitem[Yang et~al.(2024)Yang, Teng, Zheng, Ding, Huang, Xu, Yang, Hong, Zhang, Feng, et~al.]{yang2024cogvideox}
Zhuoyi Yang, Jiayan Teng, Wendi Zheng, Ming Ding, Shiyu Huang, Jiazheng Xu, Yuanming Yang, Wenyi Hong, Xiaohan Zhang, Guanyu Feng, et~al.
\newblock {CogVideoX}: Text-to-video diffusion models with an expert transformer.
\newblock \emph{arXiv preprint arXiv:2408.06072}, 2024.

\bibitem[Ye et~al.(2024)Ye, Liu, Sun, Zhou, Zhan, and Qiu]{ye2024datamixing}
Jiasheng Ye, Peiju Liu, Tianxiang Sun, Yunhua Zhou, Jun Zhan, and Xipeng Qiu.
\newblock Data mixing laws: Optimizing data mixtures by predicting language modeling performance.
\newblock \emph{arXiv preprint arXiv:2403.16952}, 2024.

\bibitem[Zheng et~al.(2024)Zheng, Peng, Yang, Shen, Li, Liu, Zhao, Li, and You]{zheng2024opensora}
Zangwei Zheng, Xiangyu Peng, Tianji Yang, Chenhui Shen, Shenggui Li, Hongxin Liu, Yukun Zhao, Yining Li, and Yang You.
\newblock Open-{Sora}: Democratizing efficient video production for all.
\newblock \emph{arXiv preprint arXiv:2412.20404}, 2024.

\end{thebibliography}

\clearpage
\newpage
\beginappendix

\FloatBarrier
\section{Attribute Specifications}
\label{app:attributes}

\tabref{tab:attributes} lists all attributes used in the Moving Alphabet dataset and their possible values.

\begin{table}[!htbp]
\centering
\caption{%
Moving Alphabet attributes and their value sets.
All attributes are sampled independently and uniformly from the listed values.
}
\label{tab:attributes}

\vspace{0.3em}
\small

\begin{tabular}{@{}llp{8cm}@{}}
\toprule
Category & Attribute & Values \\
\midrule
\multirow{5}{*}{Appearance}
& Letter identity & 52 classes (26 uppercase A--Z, 26 lowercase a--z) \\
& Font & 8 typefaces (serif, sans-serif, monospace variants) \\
& Color & 10 options (red, orange, yellow, green, cyan, blue, purple, magenta, white, black) \\
& Size & 3 levels: small (32\,px), medium (56\,px), large (80\,px) \\
& Rotation & 8 angles (0°, 45°, 90°, 135°, 180°, 225°, 270°, 315°) \\
\midrule
\multirow{2}{*}{Motion}
& Direction & Continuous angle 0--360° (binned to 4 cardinal quadrants for evaluation) \\
& Speed & Continuous range 3--10 pixels per frame (sampled uniformly) \\
\bottomrule
\end{tabular}

\vspace{-0.5em}
\end{table}

\FloatBarrier
\section{Training Details}
\label{app:training-details}

All models are trained using AdamW with a learning rate of $5\times10^{-4}$, linearly warmed up from $0$ over the first $500$ steps and held constant thereafter. We use a per-GPU batch size of $24$, bf$16$ mixed precision, and gradient clipping with a maximum norm of $1.0$. During training, we apply $10\%$ text dropout to enable classifier-free guidance during inference~\cite{ho2022classifier}.

\FloatBarrier
\section{Evaluation Metric Details}
\label{app:eval-details}

In all cases we first obtain the letter pixels with a simple foreground mask: a pixel is foreground if its brightest RGB channel exceeds $40/255$, which cleanly separates the letters from the near-black background. For 1L videos this mask isolates the single letter directly. For multi-letter videos, we additionally localize each letter by its ground-truth trajectory: at every frame we place a rectangular region of interest (ROI) at the letter's known position (recorded in the metadata, \S\ref{sec:data-gen}) and sized to its rendered glyph, intersect it with the foreground mask, and read that letter's attributes only from the pixels inside. Separating letters by location rather than appearance avoids cross-letter interference when their paths cross. Because each ROI is anchored to a specific ground-truth letter, each prediction is aligned to its corresponding letter without any matching step, and we average per-attribute accuracy over all letters across all videos at each complexity level. From these per-letter pixels, we measure:
\begin{itemize}
    \item \emph{Color accuracy} ($A_\text{color}$): We extract the brightest foreground pixels with intensity greater than 200, where anti-aliasing effects are negligible, compute their median RGB value, and assign the nearest ground-truth color based on Euclidean distance in the RGB space. The predicted color is correct if it matches the ground-truth color.
    \item \emph{Size accuracy} ($A_\text{size}$): we measure the maximum bounding-box dimension of the foreground region (median over the first 5 frames) and bin it into small ($<$44\,px), medium (44--68\,px), or large ($\geq$68\,px), corresponding to GT render sizes of 32, 56, and 80\,px. Correct if the bin matches GT.
    \item \emph{Direction accuracy} ($A_\text{dir}$): We compute the displacement vector between the first and last detected centroids, convert it to an angle using $\text{atan2}$, and assign it to one of four $90^\circ$ quadrants: right, down, left, or up. The predicted direction is correct if it matches the GT cardinal direction.
    \item \emph{Speed correlation} ($\rho_\text{speed}$): we compute the median per-frame centroid displacement (in pixels) as the predicted speed, then report the Pearson correlation between predicted and GT speeds across all videos in the evaluation set.
\end{itemize}

\FloatBarrier
\section{Training Convergence}
\label{app:convergence}

\begin{figure}[H]
    \centering
    \includegraphics[width=\textwidth]{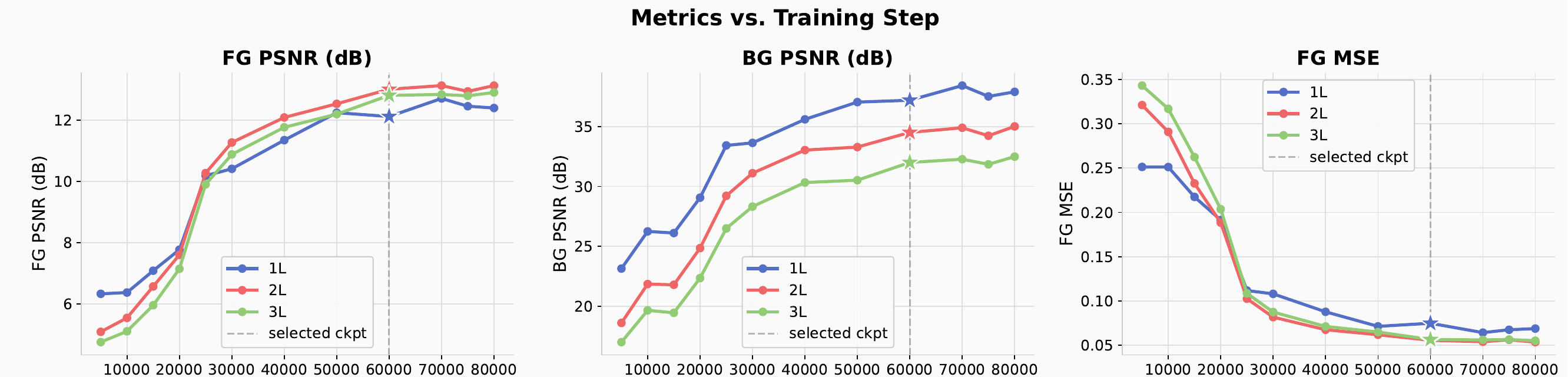}
    \caption{%
        Pixel-quality metrics (FG~PSNR, BG~PSNR, and FG~MSE) vs.\ training step for the Mix Equal condition.
        All metrics plateau by step 60K--70K.
        The dashed line marks the selected checkpoint.
    }
    \label{fig:convergence}
\end{figure}

\figref{fig:convergence} shows three pixel-quality metrics (FG~PSNR, BG~PSNR, and FG~MSE) as a function of training step for the Mix Equal condition, evaluated at two CFG values across all three complexity levels.
All metrics plateau by step 60K--70K, which corresponds to roughly 100 epochs with 150K training samples.
We select epoch 100 as the evaluation checkpoint for all experiments in this paper.

\FloatBarrier

\section{Caption Corruption: Additional Metrics}
\label{app:caption-additional}

\begin{figure}
    \centering
    \begin{subfigure}[t]{0.31\textwidth}
        \includegraphics[width=\linewidth]{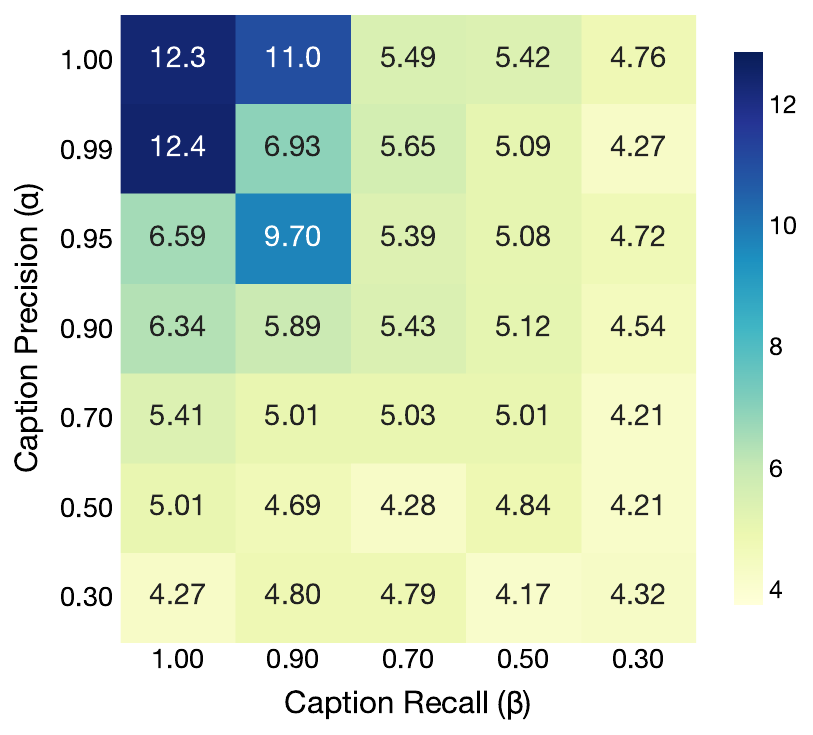}
        \caption{FG~PSNR}
        \label{fig:fg_psnr_heatmap}
    \end{subfigure}\hfill
    \begin{subfigure}[t]{0.31\textwidth}
        \includegraphics[width=\linewidth]{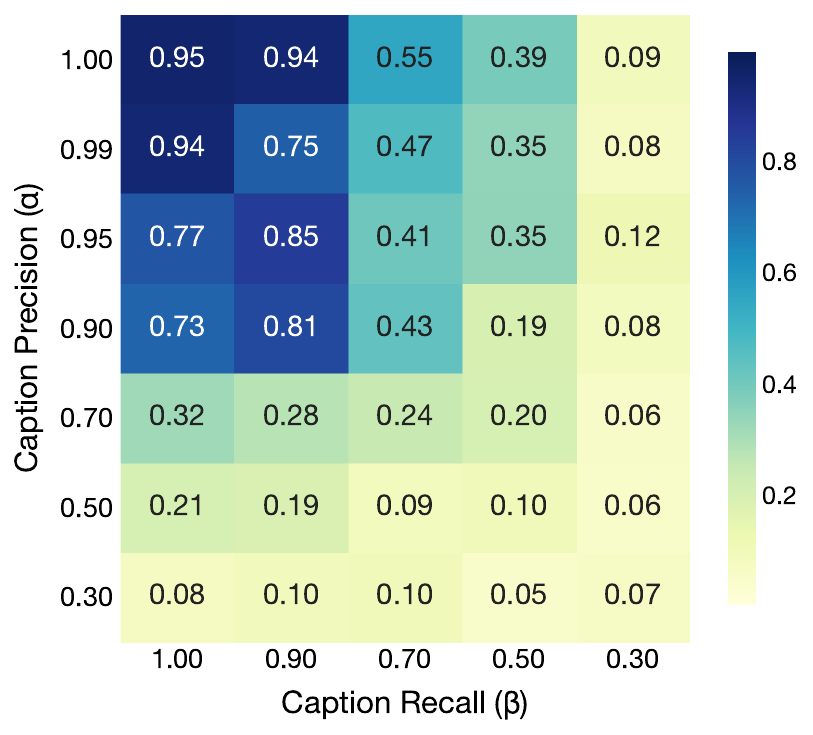}
        \caption{Color}
        \label{fig:color_heatmap}
    \end{subfigure}\hfill
    \begin{subfigure}[t]{0.31\textwidth}
        \includegraphics[width=\linewidth]{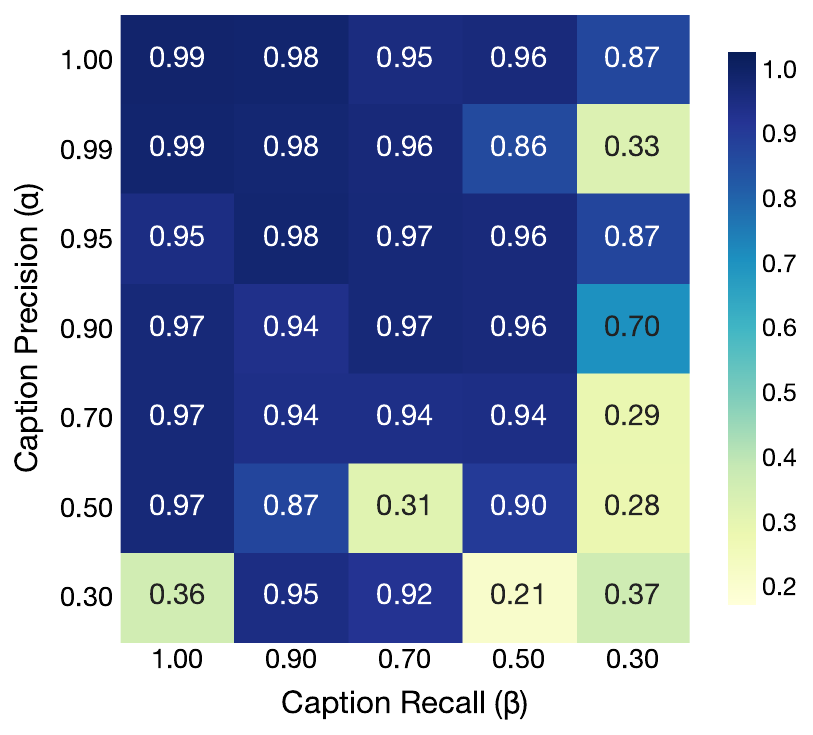}
        \caption{Direction}
        \label{fig:direction}
    \end{subfigure}

    \vspace{0.6em}
    \begin{subfigure}[t]{0.31\textwidth}
        \includegraphics[width=\linewidth]{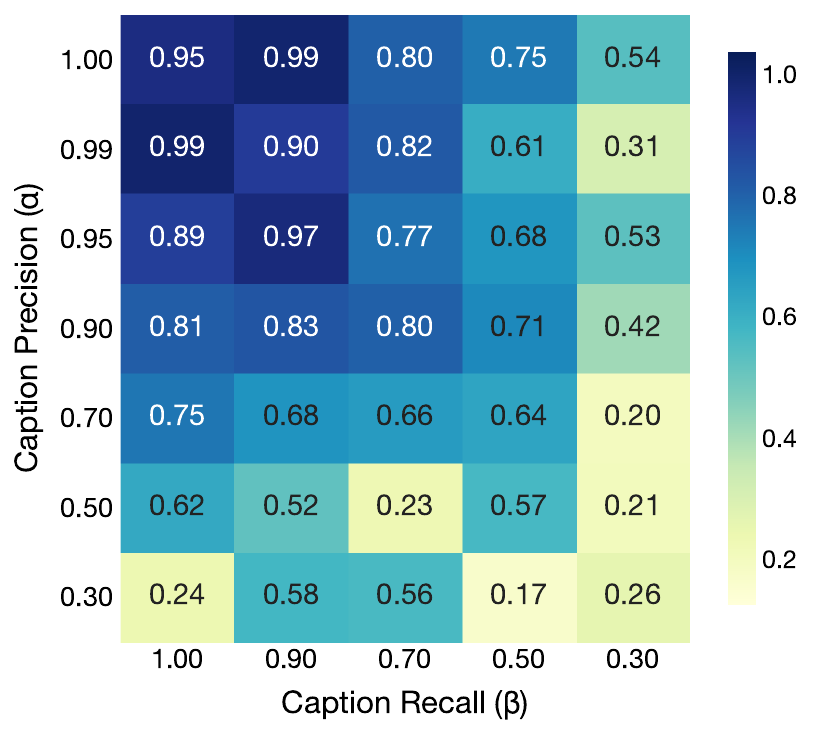}
        \caption{Size}
        \label{fig:size}
    \end{subfigure}\hfill
    \begin{subfigure}[t]{0.31\textwidth}
        \includegraphics[width=\linewidth]{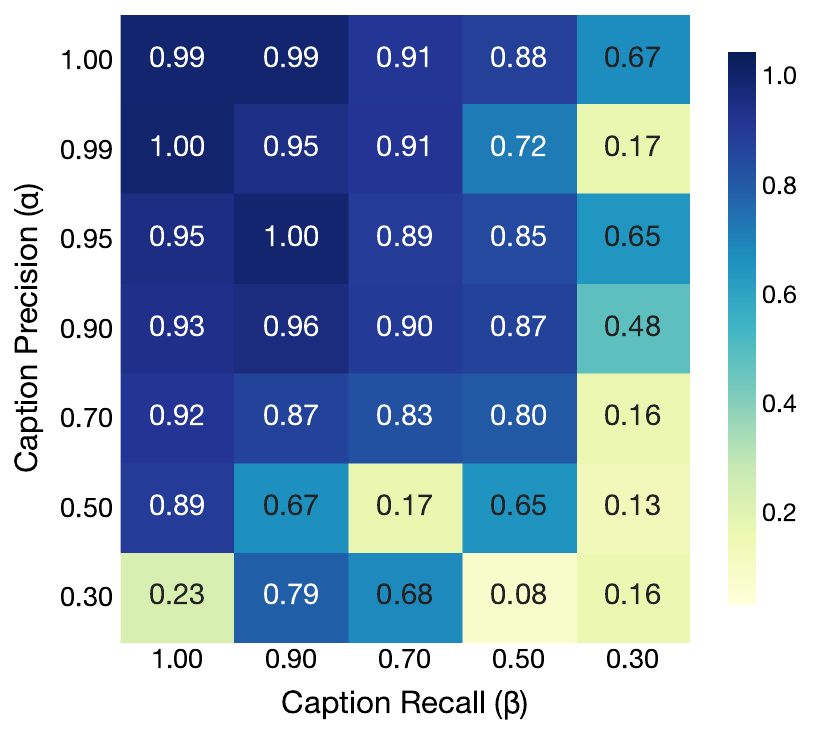}
        \caption{Speed}
        \label{fig:speed}
    \end{subfigure}\hfill
    \begin{subfigure}[t]{0.31\textwidth}~\end{subfigure}
    \caption{%
        Full $\alpha \times \beta$ heatmaps for caption precision and recall at CFG\,=\,1.0, across all metrics: FG~PSNR, color, direction, size, and speed.
        The FG~PSNR and color panels provide the per-condition detail behind the line plots in \figref{fig:precision_recall}.
    }
    \label{fig:caption_additional}
\end{figure}

\figref{fig:direction}--\ref{fig:speed} show the remaining caption corruption heatmaps from \S\ref{sec:precision-vs-recall} that are not included in the main figure.

\begin{figure}[H]
    \centering
    \includegraphics[width=\linewidth]{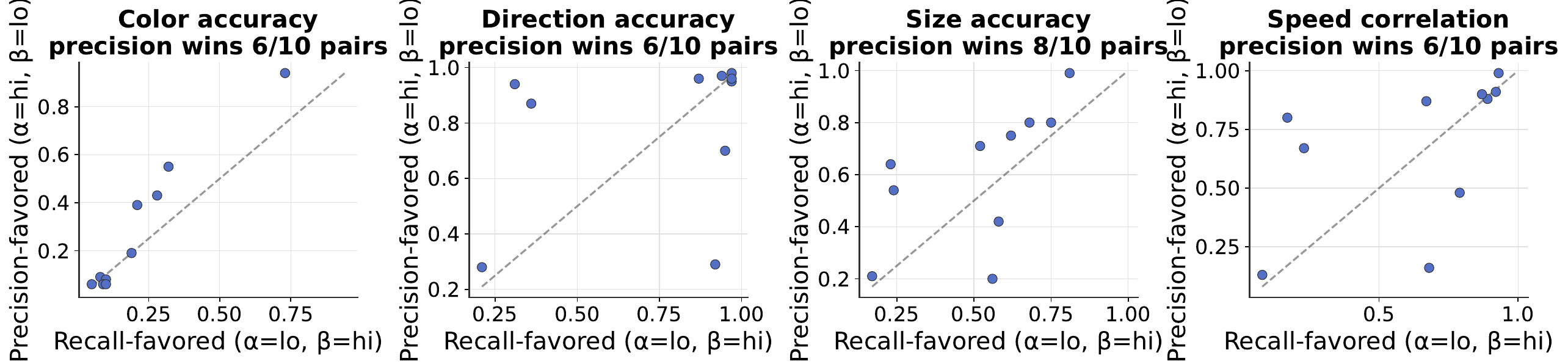}
    \caption{%
        Per-pair detail behind the precision advantage in \figref{fig:precision_recall}(c).
        For every pair of corruption levels, we plot the metric when the higher level is assigned to precision ($y$, $\alpha{=}$hi, $\beta{=}$lo) against the recall-favored assignment ($x$, $\alpha{=}$lo, $\beta{=}$hi); points above the diagonal favor precision.
        Precision-favored allocations win in the majority of pairs for every attribute, and the wins are large for color and size; direction and speed are noisier (single run, CFG\,=\,1.0).
    }
    \label{fig:swap_scatter}
\end{figure}

\figref{fig:swap_scatter} provides the per-pair detail behind the precision advantage in \figref{fig:precision_advantage}(c): for every pair of corruption levels, we compare assigning the higher level to precision against assigning it to recall. Precision-favored allocations win for the majority of pairs across all four attributes, with the largest margins for color and size.

\section{Per-Attribute Breakdowns}
\label{app:per-attr}

\begin{figure}[H]
    \centering
    \includegraphics[width=0.32\textwidth]{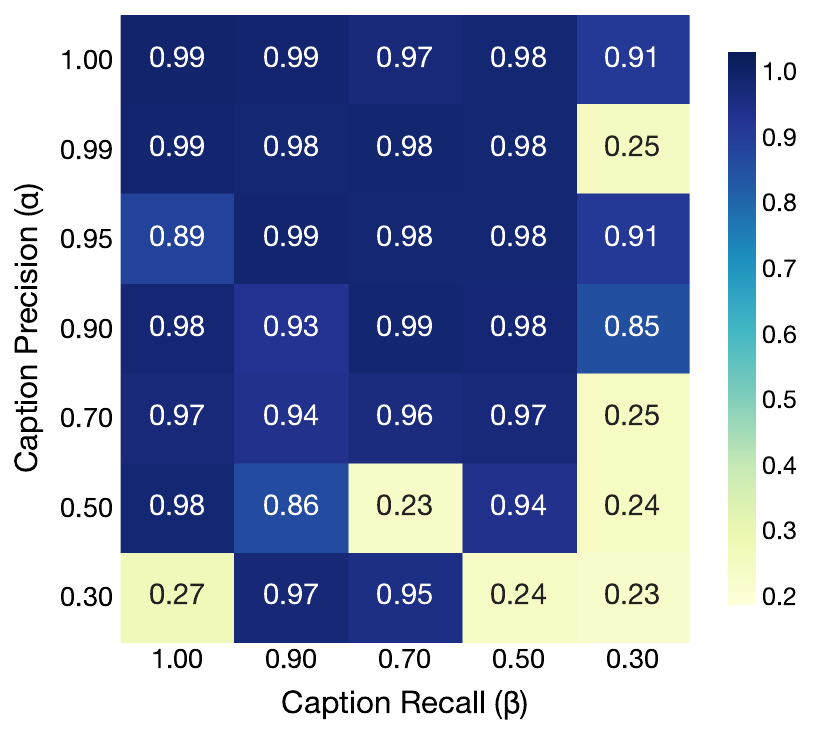}
    \includegraphics[width=0.32\textwidth]{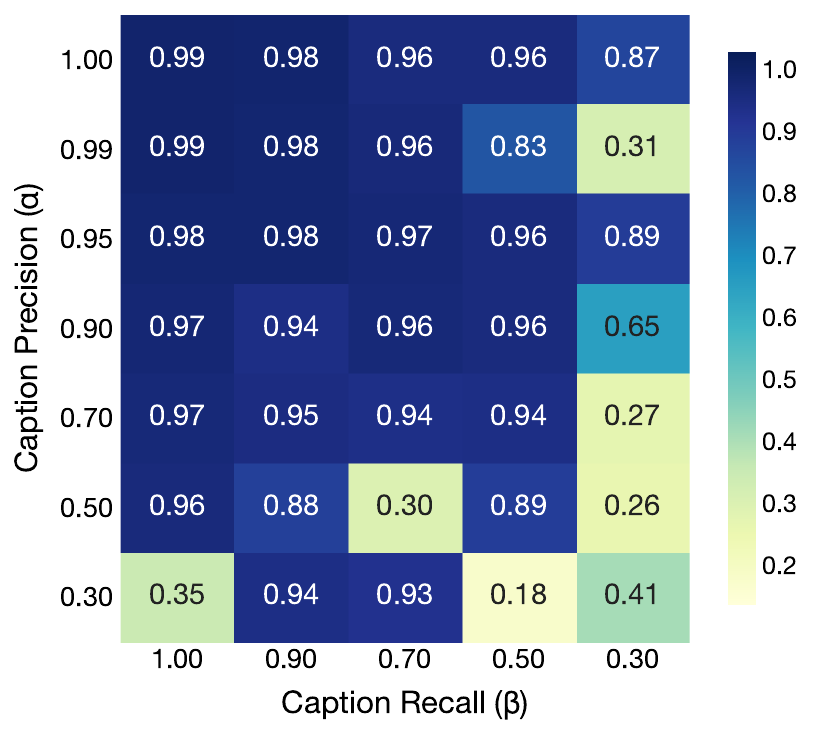}
    \includegraphics[width=0.32\textwidth]{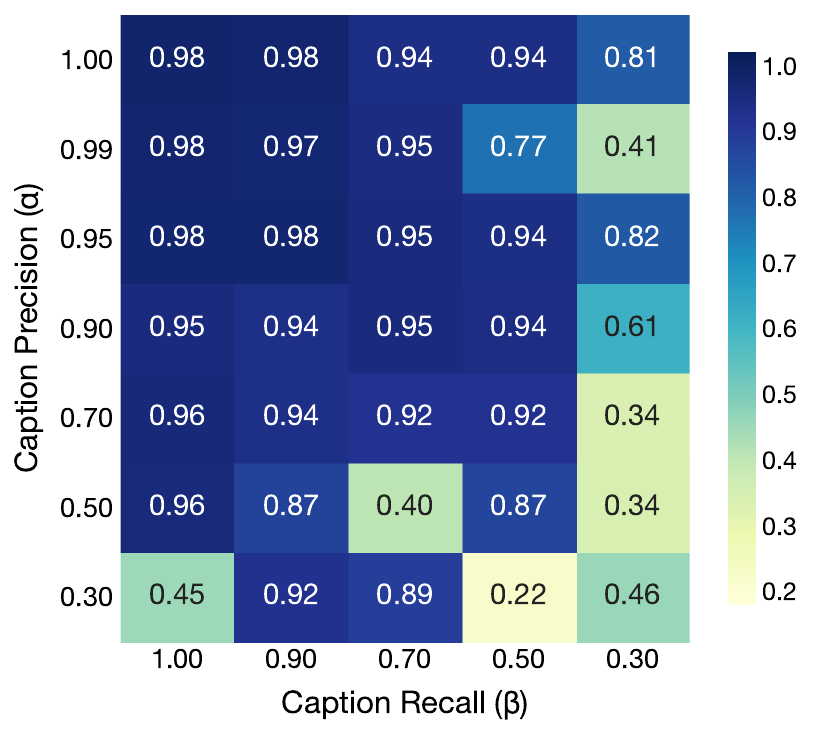}
    \caption{Direction accuracy across the $\alpha \times \beta$ caption corruption grid, evaluated separately at 1L (left), 2L (center), and 3L (right) complexity. CFG\,=\,1.0.}
    \label{fig:per-attr-direction}
\end{figure}

\begin{figure}[H]
    \centering
    \includegraphics[width=0.32\textwidth]{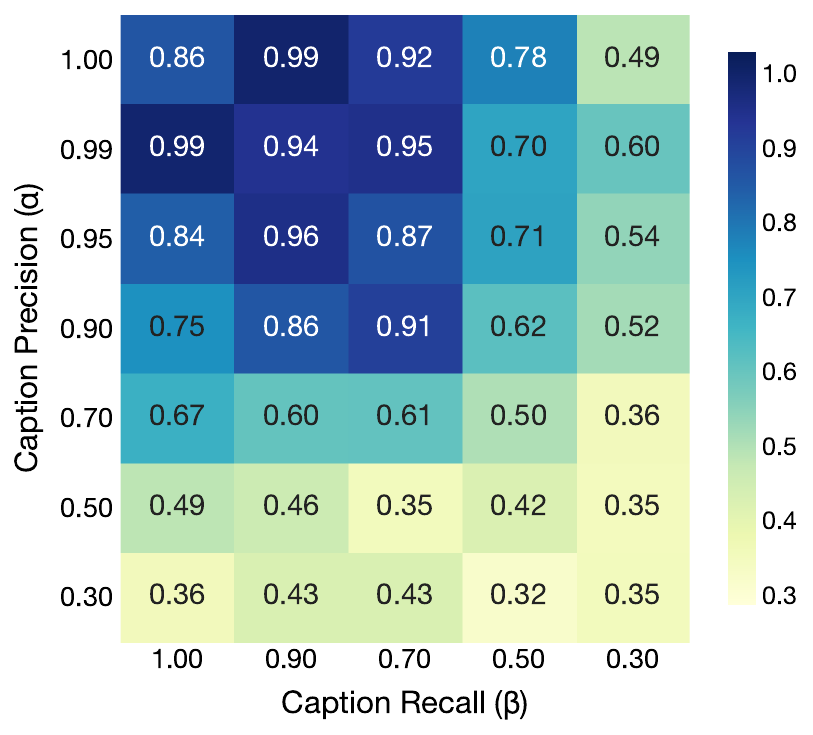}
    \includegraphics[width=0.32\textwidth]{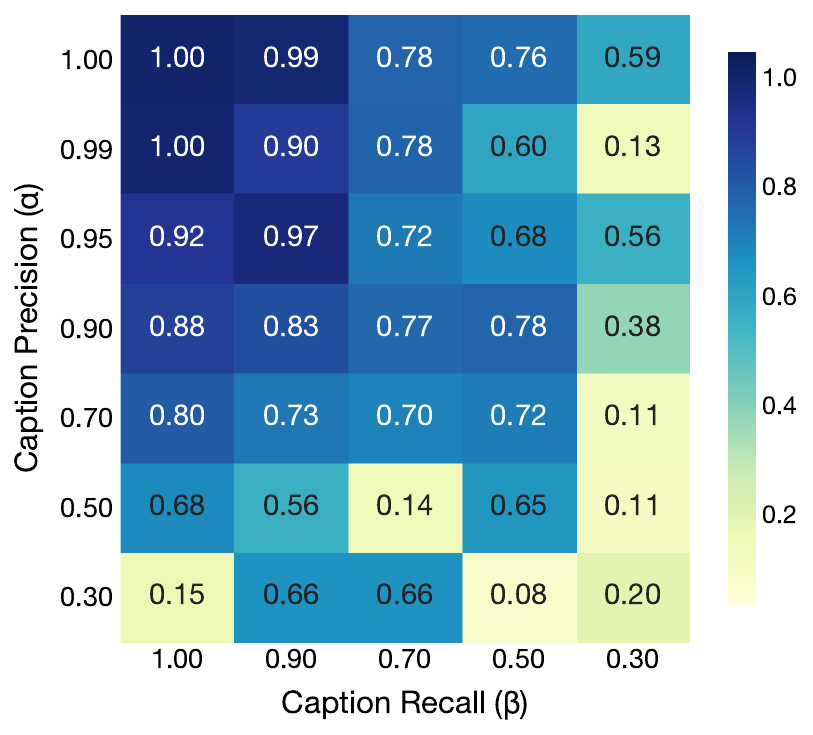}
    \includegraphics[width=0.32\textwidth]{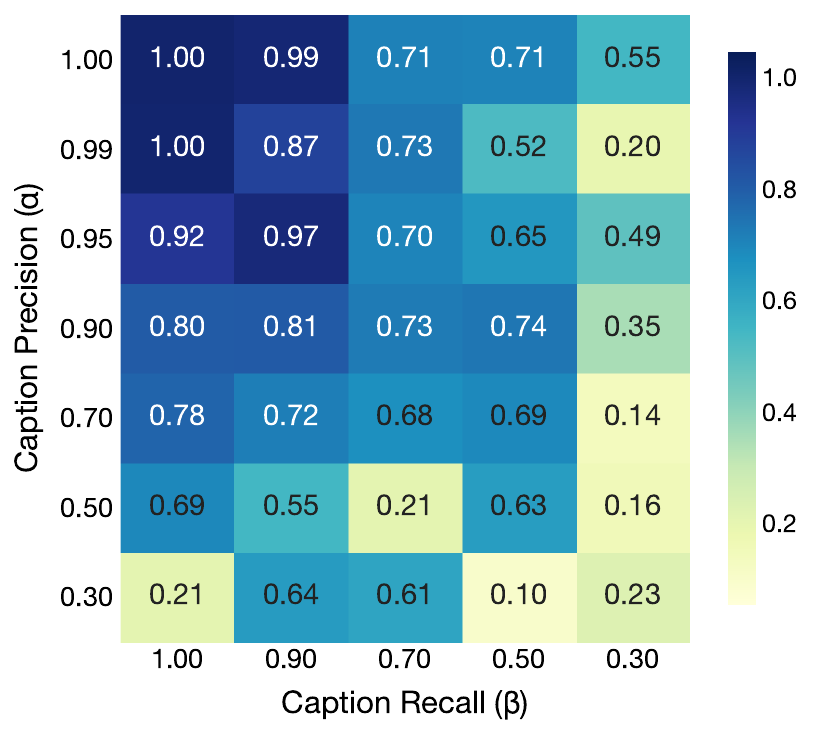}
    \caption{Size accuracy across the $\alpha \times \beta$ grid at 1L, 2L, and 3L complexity. CFG\,=\,1.0.}
    \label{fig:per-attr-size}
\end{figure}

\begin{figure}[H]
    \centering
    \includegraphics[width=0.32\textwidth]{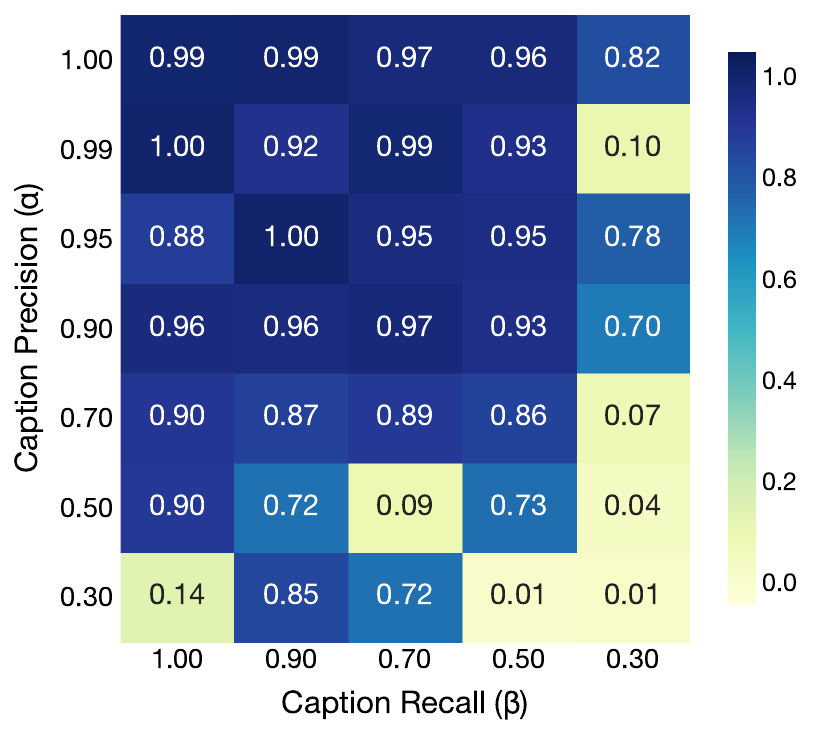}
    \includegraphics[width=0.32\textwidth]{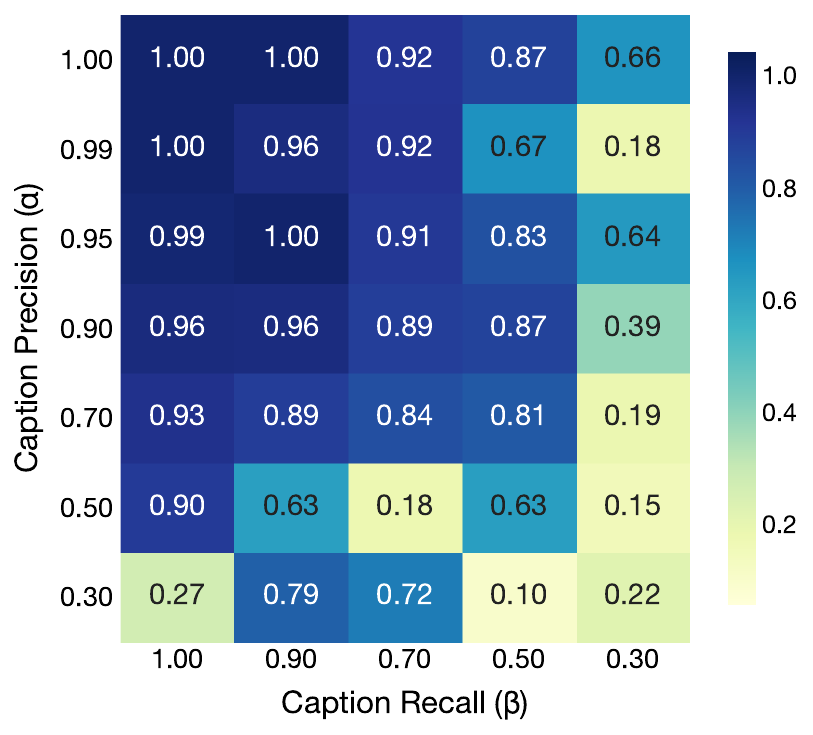}
    \includegraphics[width=0.32\textwidth]{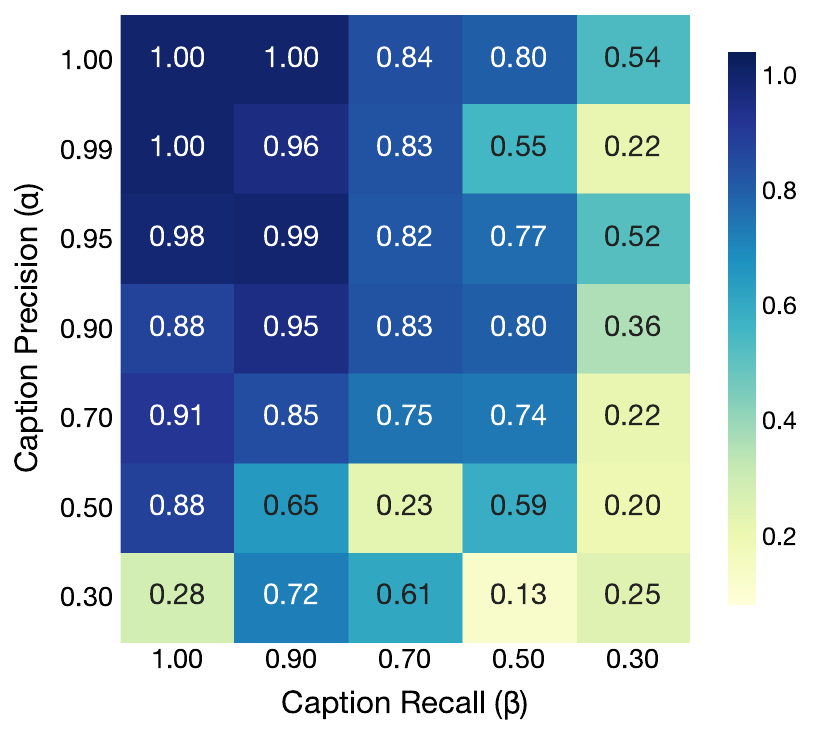}
    \caption{Speed correlation across the $\alpha \times \beta$ grid at 1L, 2L, and 3L complexity. CFG\,=\,1.0.}
    \label{fig:per-attr-speed}
\end{figure}

\begin{figure}[H]
    \centering
    \includegraphics[width=0.32\textwidth]{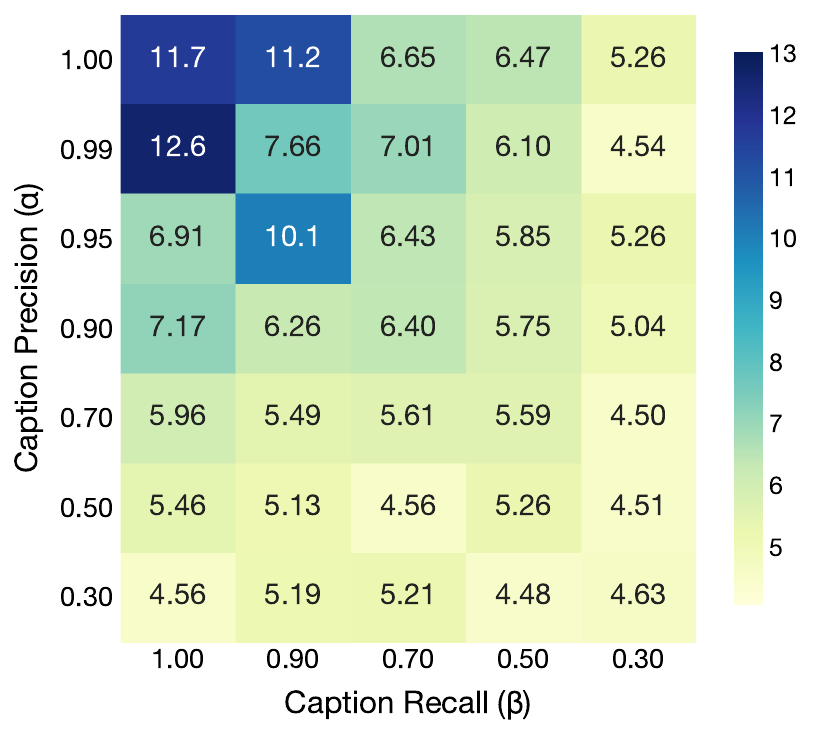}
    \includegraphics[width=0.32\textwidth]{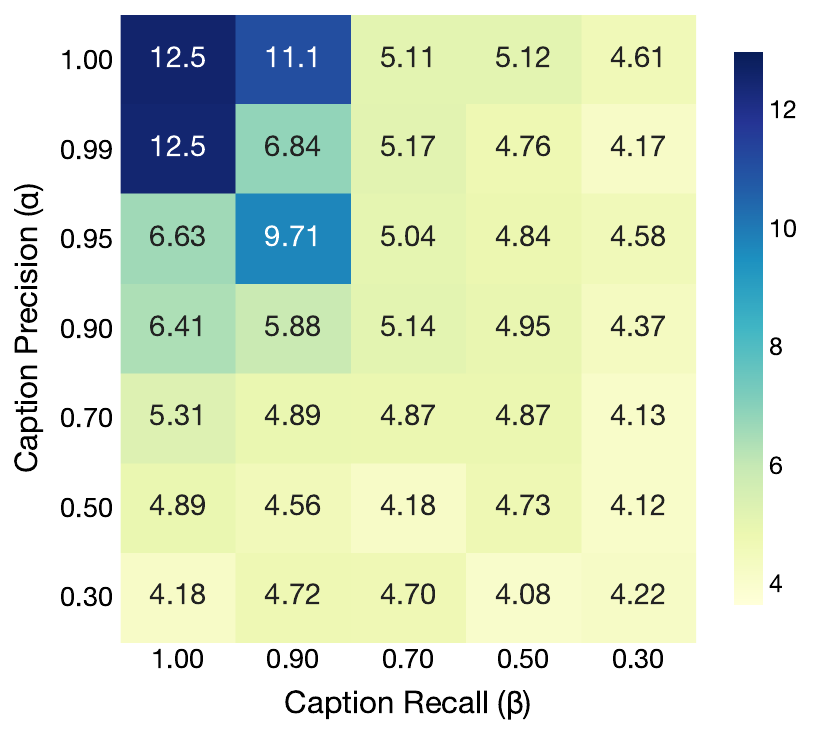}
    \includegraphics[width=0.32\textwidth]{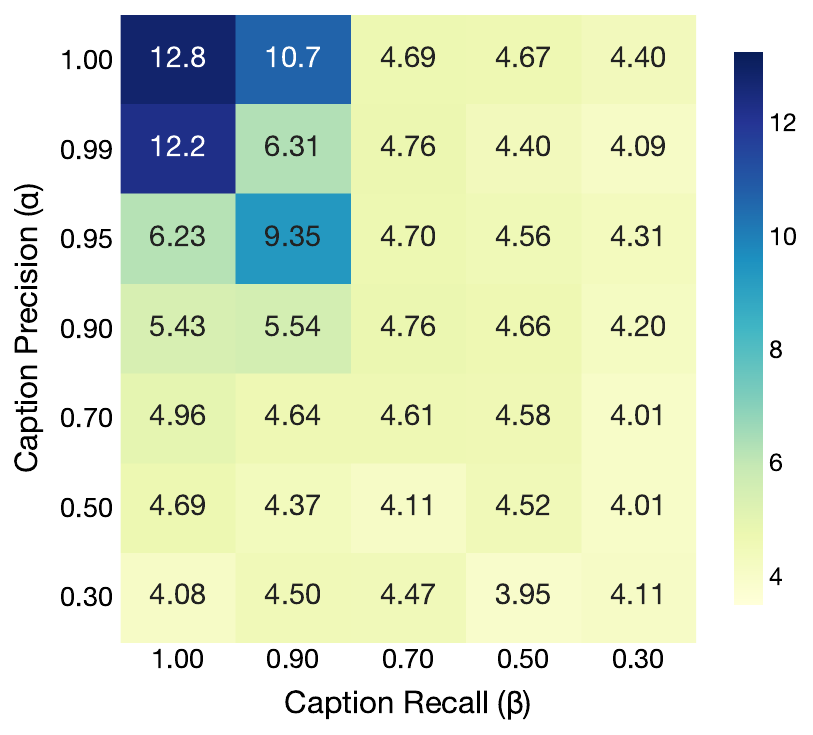}
    \caption{Foreground PSNR across the $\alpha \times \beta$ grid at 1L, 2L, and 3L complexity. CFG\,=\,1.0.}
    \label{fig:per-attr-fg-psnr}
\end{figure}

\figref{fig:per-attr-direction}--\ref{fig:per-attr-fg-psnr} show per-level heatmaps for the caption corruption experiment from \S\ref{sec:caption-quality}, breaking down each metric by complexity level (1L, 2L, 3L) across the $\alpha \times \beta$ grid at CFG\,=\,1.0.
At higher complexity, models degrade more steeply with caption corruption, and the gap between $\alpha$ and $\beta$ effects becomes more pronounced.

\FloatBarrier

\section{Marginal Bias: Single-Attribute Details}
\label{app:marginal-attr}

\begin{figure}[H]
    \centering
    \includegraphics[width=\linewidth]{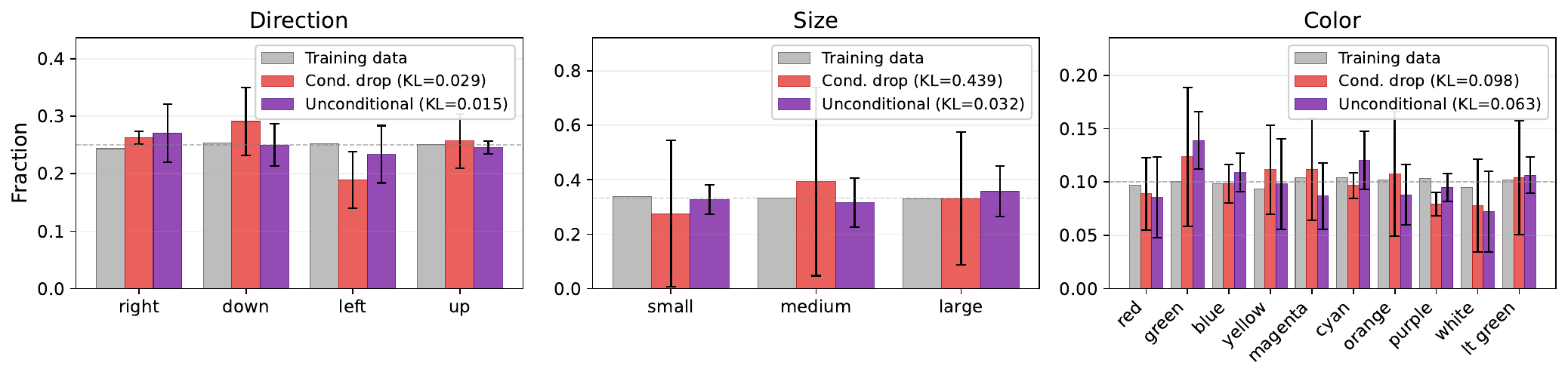}
    \caption{%
        Single-attribute setting: GT vs.\ conditional drop vs.\ unconditional for each attribute.
        Bars show mean over 4 seeds; error bars show $\pm$1 std.
        The conditional drop model develops large, high-variance biases for size, while the unconditional model stays near-uniform.
    }
    \label{fig:implicit_defaults}
\end{figure}

\begin{figure}[H]
    \centering
    \includegraphics[width=0.55\linewidth]{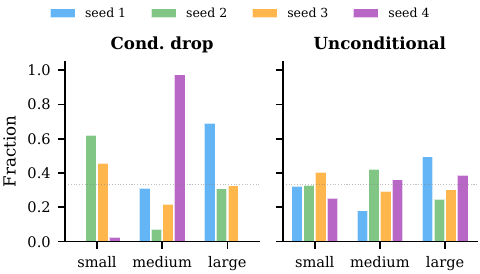}
    \caption{%
        Single-attribute setting: per-seed size distributions for conditional drop vs.\ unconditional models.
        Each conditional drop seed converges to a different dominant size, while unconditional models preserve diversity.
    }
    \label{fig:size_per_seed}
\end{figure}

\begin{table}[H]
\centering
\caption{%
    KL divergence from the training distribution when an attribute is never mentioned in captions.
    \textbf{Cond.\ drop}: text-conditioned model with the attribute omitted from all captions.
    \textbf{Unconditional}: model trained without any text conditioning.
    Values are mean $\pm$ std over 4 random seeds (CFG\,=\,1.0).
}
\label{tab:marginal_attr}
\vspace{0.3em}
\small
\setlength{\tabcolsep}{5pt}
\begin{tabular}{@{}lccc@{}}
\toprule
Attribute & Cond.\ drop & Unconditional & Ratio \\
\midrule
Size      & $0.44 \pm 0.35$ & $0.03 \pm 0.03$ & $13.7\times$ \\
Color     & $0.10 \pm 0.04$ & $0.06 \pm 0.04$ & $1.6\times$  \\
Direction & $0.03 \pm 0.03$ & $0.01 \pm 0.01$ & $1.9\times$  \\
\bottomrule
\end{tabular}
\vspace{-0.5em}
\end{table}

In the more controlled single-attribute setting, the collapse is far more extreme.
\tabref{tab:marginal_attr} shows that conditional drop models have $13.7\times$ higher KL divergence than unconditional models for size, and \figref{fig:size_per_seed} reveals the mechanism: each seed collapses to a different dominant size, while all unconditional models maintain roughly uniform distributions.
The text encoder's null representation for size acts as a strong but arbitrary default that varies with random initialization.
For color ($1.6\times$) and direction ($1.9\times$), the gap between conditional drop and unconditional is small, and both conditions show similar bias patterns (\figref{fig:implicit_defaults}).

\end{document}